\documentclass[a4paper,fleqn]{cas-dc}

\usepackage[capitalize]{cleveref}
\usepackage{graphicx}
\usepackage{caption}
\usepackage{subfigure}
\usepackage{float}
\usepackage{stfloats}
\hypersetup{linkcolor=blue,
            anchorcolor=blue,
            citecolor=blue}

\usepackage[authoryear]{natbib}
\def\tsc#1{\csdef{#1}{\textsc{\lowercase{#1}}\xspace}}
\tsc{WGM}
\tsc{QE}
\begin{document}\sloppy
\captionsetup[figure]{labelfont={bf},name={Fig.},labelsep=period}
\let\WriteBookmarks\relax
\def\floatpagepagefraction{1}
\def\textpagefraction{.001}
\let\printorcid\relax
% Short title
\shorttitle{Motion Detection Inspired by Birds}    

% Short author
\shortauthors{Pingge Hu et al.}  

% Main title of the paper
\title [mode = title]{TSOM: Small Object Motion Detection Neural Network Inspired by Avian Visual Circuit}  

% Title footnote mark
% eg: \tnotemark[1]
% \tnotemark[<tnote number>] 

% Title footnote 1.
% eg: \tnotetext[1]{Title footnote text}
% \tnotetext[<tnote number>]{<tnote text>} 

% First author
%
% Options: Use if required
% eg: \author[1,3]{Author Name}[type=editor,
%       style=chinese,
%       auid=000,
%       bioid=1,
%       prefix=Sir,
%       orcid=0000-0000-0000-0000,
%       facebook=<facebook id>,
%       twitter=<twitter id>,
%       linkedin=<linkedin id>,
%       gplus=<gplus id>]

\author[1]{Pingge Hu}\fnmark[1]
\ead{hpg18@mails.tsinghua.edu.cn}
\author[1]{Xiaoteng Zhang}\fnmark[1]
\ead{zxt21@mails.tsinghua.edu.cn}
\author[2,3]{Mengmeng Li}
\author[4]{Yingjie Zhu}\cormark[1]
\author[1,3]{Li Shi}\cormark[1]
\ead{shilits@tsinghua.edu.cn}

% Corresponding author indication
\cortext[1]{Corresponding author.}

\affiliation[1]{organization={Department of Automation, Tsinghua University},
            city={Beijing},
            postcode={100084},
            country={China}}
\affiliation[2]{organization={School of Electrical and Information Engineering, Zhengzhou University},
            city={Zhengzhou},
            postcode={450001},
            country={China}}
\affiliation[3]{organization={Henan Key Laboratory of Brain Science and Brain-Computer Interface Technology},
            city={Zhengzhou},
            postcode={450001},
            country={China}}
\affiliation[4]{organization={Shenzhen Key Laboratory of Drug Addiction, Shenzhen Neher Neural Plasticity Laboratory, Brain Cognition and Brain Disease Institute, Shenzhen Institute of Advanced Technology, Chinese Academy of Sciences},
            city={Shenzhen},
            postcode={518055},
            country={China}}
% Footnote of the first author
\fntext[1]{These authors contribute equally to this work.}

% URL of the first author
% \ead[url]{<URL>}

% Credit authorship
% eg: \credit{Conceptualization of this study, Methodology, Software}
% \credit{<Credit authorship details>}

% % Address/affiliation
% \affiliation[1]{organization={Department of Automation, Tsinghua University},
%             city={Beijing},
%             postcode={100084},
%             country={China}}

% \author[<aff no>]{<author name>}[<options>]

% % Footnote of the second author
% \fnmark[2]

% % Email id of the second author
% \ead{}

% % URL of the second author
% \ead[url]{}

% % Credit authorship
% \credit{}

% % Address/affiliation
% \affiliation[<aff no>]{organization={},
%             addressline={}, 
%             city={},
% %          citysep={}, % Uncomment if no comma needed between city and postcode
%             postcode={}, 
%             state={},
%             country={}}

% % Corresponding author text
% \cortext[1]{Corresponding author}

% % Footnote text
% \fntext[1]{}

% For a title note without a number/mark
%\nonumnote{}

% Here goes the abstract
\begin{abstract}
%% Text of abstract
Detecting small moving objects in complex backgrounds from an overhead perspective is a highly challenging task for machine vision systems. As an inspiration from nature, the avian visual system is capable of processing motion information in various complex aerial scenes, and its Retina-OT-Rt visual circuit is highly sensitive to capturing the motion information of small objects from high altitudes. However, more needs to be done on small object motion detection algorithms based on the avian visual system. In this paper, we conducted mathematical modeling based on extensive studies of the biological mechanisms of the Retina-OT-Rt visual circuit. Based on this, we proposed a novel tectum small object motion detection neural network (TSOM). The neural network includes the retina, SGC dendritic, SGC Soma, and Rt layers, each layer corresponding to neurons in the visual pathway. The Retina layer is responsible for accurately projecting input content, the SGC dendritic layer perceives and encodes spatial-temporal information, the SGC Soma layer computes complex motion information and extracts small objects, and the Rt layer integrates and decodes motion information from multiple directions to determine the position of small objects. Extensive experiments on pigeon neurophysiological experiments and image sequence data showed that the TSOM is biologically interpretable and effective in extracting reliable small object motion features from complex high-altitude backgrounds.

\end{abstract}

% Use if graphical abstract is present
%\begin{graphicalabstract}
%\includegraphics{}
%\end{graphicalabstract}

% Research highlights
\begin{highlights}
\item Mechanism of Retina-OT-Rt neural circuit is mathematically modeled and used for small object motion detection.
\item A novel neural network called TSOM is proposed for detecting small object motion.
\item TSOM model has high biological interpretability and can simulate the response of neurons.
\item TSOM model has better detection performance than other advanced methods.
\end{highlights}

% Keywords
% Each keyword is seperated by \sep

\maketitle

% Main text
\section{Introduction}\label{Introduction}

Small object detection is critical in various complex application scenarios, such as Unmanned Aerial Vehicle (UAV) scene analysis, remote sensing telemetry, and surveillance. In object detection, regular-sized objects have been able to be detected accurately. In contrast, small object detection is still a challenging problem, especially when it comes to motion detection of small objects in high-altitude scenes. Specifically, small objects occupy few pixel points and low resolution in the image, resulting in few compelling features for small objects. At the same time, the background in high-altitude scenes is cluttered, making it easy to confuse small objects with the background. The above difficulties become challenging when faced with a situation where the small object has a relative motion with the background.

Over the past decades, scholars have conducted much research in small object motion detection. Conventional algorithms, such as optical flow \citep{liu2023exploring}, frame difference \citep{wang2021moving}, and background subtraction \citep{kalsotra2022background,garcia2020background} methods, have made significant progress but exhibit insufficient robustness when dealing with scenes with relative motion between small objects and the background. Meanwhile, deep learning methods, such as Convolution Neural Networks (CNN) \citep{tezcan2021bsuv,min2022attentional} and Transformer \citep{liu2023infrared,liu2024msrmnet}, have achieved great success in many fields but are still limited in their ability to extract valid information in objects with only very few pixel points.

The brain shows extraordinary abilities in perceiving the exterior environment in the natural world. Therefore, building a system mimicking a biological brain for visual perception has always been attractive. The biological visual system receives considerable information and employs distinct information processing techniques when dealing with complex scenarios. Flying animals exhibit exceptional sensitivity to motion due to their ability to maneuver in the vertical dimension freely. In situations of high altitude, birds exhibit superior visual abilities. As a result, the excellent small object motion detection ability of birds in high-altitude scenes can provide new insights for current artificial intelligence algorithms.

Birds have three main visual pathways: the tectofugal pathway, the thalamofugal pathway, and the accessory optic system \citep{ref1}. Behavioral investigations have suggested that the tectofugal pathway is more significant than other pathways for recognizing patterns and fine visual discrimination in birds. Thus, the tectofugal pathway is the foremost visual pathway in birds, accountable for generating spatial perception, motion detection, and attention control \citep{ref2}. The Optic Tectum (OT) is a fundamental element of the avian visual pathway that gathers information about stimuli position and computes saliency in the environment \citep{ref3,ref4}. OT sensory neurons are driven best by discrete, space-specific stimuli \citep{stein1993merging}, which has been reported in a wide range of species, including goldfish \citep{schellart1979center}, pigeons \citep{frost1981moving}, owls \citep{ref7}, and monkeys \citep{munoz1998lateral}. Compared to other animal species, the avian OT is relatively large, comprising 15 layers \citep{ref5}. Josine et al. \citep{ref6} suggest that OT typically reacts most to small white stimuli. Mysore et al. \citep{ref7} confirm that OT exhibits a heightened response to small, rapidly flickering stimuli. The OT-centered Retina-OT-Rt (Retina-OT-Rotundal) neural circuit is responsible for perceiving and extracting small object motion, indicating the avian aptitude for precise spatial localization of prey.

Motivated by the superior properties of birds, researchers have started to model the Retina-OT-Rt neural circuit, aiming to apply it to small object detection. Wang et al. \citep{ref16} proposed the energy accumulation model for the dendritic field of Stratum Griseum Centrales (SGCs), which models the sensitivity of SGCs to continuous motion preference \citep{ref24}. However, their study is limited to single-layer neural modeling without considering the modeling of visual neural circuits. As a result, it cannot be applied to real-world small object detection tasks. Dellen et al. \citep{ref29} modeled the information processing of the Retina-OT-Rt circuit as a global Fourier transform for velocity estimation. Huang et al. \citep{ref31} proposed an object motion detector based on the elementary motion detector by modeling the accumulation properties of OT neurons. However, both modeling studies ignore the circuit's sensitivity to the small-scale object features.

To overcome these limitations, this paper introduces a small object motion neural network based on the avian visual circuit. Our contributions are as follows:
 
\begin{enumerate}[1)]
\item To dissect the Retina-OT-Rt neural mechanisms, we build the mathematical description of the Retina-OT-Rt neural circuit.
\item To extract efficient features of small objects in the real-world image sequence, we propose a neural network called TSOM (Tectum Small Object Motion detector) based on the mathematical description for detecting the motion of small objects according to the mathematical description.
\item We verified that TSOM has biological consistency and superior performance for small object extraction over image sequences through neurophysiological and algorithm performance experiments.
\end{enumerate}

The rest of the paper is organized as follows. Section \ref{Preliminaries} describes the basics of the properties of the components and their connections to each other of the Retina-OT-Rt neural circuit. Section \ref{Related Works} presents related work of modeling the Retina-OT-Rt neural circuit. Section \ref{Methodology} introduces the proposed neural network model, including the mathematical description of the Retina-Optic-Rt information processing framework and the TSOM neural network model corresponding to the mathematical description. Section \ref{Experiment and Results} employs neurophysiology data and image sequences data to evaluate the biological properties and effectiveness in small moving object extraction of the TSOM. Section \ref{Conclusion} summarizes the paper, and future research directions are envisaged.

\section{Preliminaries}\label{Preliminaries}
The Retina-OT-Rt neural circuit plays a vital role in the feature extraction of small moving objects, essential for calculating visual saliency within the avian visual pathway. Consequently, researchers have carried out considerable anatomical, immunohistochemical, electrophysiological, and behavioral investigations on this neural pathway.

The Retina-OT-Rt neural circuit is composed of retinal ganglion cells (RGCs) located in the retina, neurons located in the central layer (Stratum Griseum Centrale, SGC) of the optic tectum (OT), and rotunda (Rt) neurons. The OT acts as the primary nucleus for receiving projections from RGCs, which transmit external stimuli into the brain. Gonzalo et al. \citep{ref8} confirm that SGC neurons found in the 13th layer of the OT receive input from RGCs and topographically project to the Rt, serving as the core connection of the entire pathway.

SGC neurons have several distinctive features. Luksch et al. \citep{ref9} carried out a study that identified three types of SGC neurons: SGC-I, SGC-II, and SGC-III. SGC-I neurons possess dendritic terminals that end in layer 5b of the OT, receiving input directly from RGCs. Whenever SGC neurons are mentioned here, they refer to SGC-I neurons.

One noteworthy attribute of SGC neurons is their vast dendritic span. While standard neurons have dendritic spans ranging from 1200 to 2000 µm, SGC neurons can have horizontal end-to-end ranges of up to 4000 µm \citep{ref9}. This characteristic provides SGC neurons with vast receptive fields, allowing them to receive small, moving stimuli and background information at the same time \citep{luksch2001chattering}. Additionally, SGC neurons have unique brush-like dendritic terminals. The RGC terminal structures in the SGC 5b layer exhibit diameters of less than 1 µm and comprise multiple radial protrusions possessing multiple synaptic orifices. Extensive overlap of these structures with SGC dendrites facilitates robust input to a brush-like dendritic terminal, enabling connectivity with either a single large or multiple small RGCs \citep{ref9}. This indicates that a sole input from a small RGC is ample for SGC neuron stimulation \citep{mahani2006sparse}.

Thus, the receptive fields of deep parietal neurons have relatively small excitatory centers and large peripheral inhibitory areas. Experiments \citep{ref8} have shown that deep parietal neurons respond to the relative motion between the test stimulus and the background rather than to the absolute direction of the stimulus.

The Rt is the most prominent single nucleus in the thalamus of most birds and has several anatomical branches, each receiving projections from a different subpopulation of SGC neurons. A distinctive feature of SGC-Rt projections is the complete loss of one-to-one correspondence \citep{ref8}. That is, the large receptive fields of SGC neurons coarsen the precise spatial map of the OT surface at the level of the SGC, and the SGC appears to rearrange the projections on the Rt more radically.

An experimental study of the form and nature of RGC-SGC-Rt projections by Gonzalo et al. \citep{ref8} in 2003 showed that SGC neurons are present over the entire parietal surface so that the population of SGC neurons retains information over the entire input visual field. For SGC-Rt, retrograde tracking experiments suggested that the projection pattern is an interlaced projection, where axons from a sparsely distributed population of neurons across the parietal surface converge on a subregion, and adjacent regions within the same subregion receive information related to a different but mixed set of SGC neurons of the same class. In this process, the precise topological map projected onto the retina is transformed into a functional map.

Due to the suitable neural response mechanism of the Retina-OT-Rt circuit for small object motion detection, this paper attempts to mathematically model the mechanism for use in small object motion detection.

\section{Related Works}\label{Related Works}
Over the years, researchers have proposed various biological neural computational models to simulate the Retina-OT-Rt neural circuit, aiming to understand better the mechanisms by which birds extract and select salient objects.

The Retina-OT-Rt neural circuit, also known as the RGC-OT-Rt neural circuit, exhibits a solid response to small moving stimuli. Dellen et al. \citep{ref29} proposed the Dellen model, a neural network model for the RGC-OT-Rt circuit, which simulates the motion-sensitive characteristics of OT and explores the organization of spatial information in this pathway. The Dellen model is based on the motion constraint equations \citep{ref30} inspired by human vision. The method utilizes the concept that the motion of a translating two-dimensional image corresponds to feature planes defined in the Fourier domain through motion constraint equations, which link image velocity to the spatiotemporal frequencies in the Fourier space.

The SGC neurons in the optic tectum of birds possess a wide-field architecture and demonstrate robust responses to small-scale stimuli and rapid motion. Wang et al. \citep{ref16} coined the term 'continuous motion-sensitive neurons' for these neurons, which are characterized by individual dendrites capable of inducing solid somatic responses with local inhibition by horizontal cells, drawing inspiration from STMD. They proposed the Directional Energy Accumulation Model to simulate the elicitation of neuron responses by individual dendrites. The effectiveness of this model was qualitatively validated through electrophysiological experiments. Huang et al. \citep{ref31} developed the EMD\_TSADM model, which is based on the ESTMD model and incorporates both the spatiotemporal accumulation coding mechanism and the dynamic surround modulation mechanism of OT neurons in the avian RGC-OT-Rt neural circuit. Their study demonstrated the biological plausibility of this model and showcased its effectiveness in detecting moving objects within natural scenes. In summary, previous research on avian visual modeling has primarily focused on constructing motion computation models for the RGC-OT-Rt neural circuit, with limited analysis of small objects within this circuit.

The research on models for detecting small object motion in insects based on Elementary Motion Detector (EMD) has been extensively studied and has significant implications for developing avian small object detection models. Insects have a relatively simple visual system that is highly flexible and efficient, allowing them to detect small moving objects in cluttered and dynamic backgrounds. Insects have cells similar to motion detectors, called Small Target Motion Detector neurons (STMD). Decades of neurophysiological research on STMD have led to the proposal of various quantitative models, such as ESTMD \citep{ref32}, DSTMD \citep{ref33}, STMD+ \citep{ref34}, and apg-STMD \citep{ref28}. These models demonstrate the ability to detect small moving targets in cluttered backgrounds. However, insights gained from insect studies may not directly contribute to improving small object motion detection algorithms in high-altitude overhead scenarios due to significant differences in living environment, visual system structure, and information processing methods between insects and birds.

\section{Methodology}\label{Methodology}
This section outlines the main contents of this paper, including the mathematical modeling of the avian visual pathway in Section \ref{Mechanism model of Retina-OT-Rt Pathway} and the mathematical description for small object motion detection established by simulating the working mechanism of the avian visual pathway in Section \ref{Retina-OT-Rt Neural Model for Small Object Motion Detection}.
\subsection{Mathematical Description of Retina-OT-Rt Circuit}\label{Mechanism model of Retina-OT-Rt Pathway}
There are two crucial unidirectional information transfer projections in the entire pathway. The initial process entails the projection from the retina to the SGC, and the second mechanism constitutes the projection from SGC axons to the Rt neurons. This paper presents a spatiotemporal dynamic mathematical depiction for SGC neuron dendritic fields in Section \ref{Spatiotemporal Dynamic Model of SGC}, as well as a two-stage motion information integration formulation at the Rt site for tracking small objects moving within the visual field in Section \ref{Two-Stage Motion Information Integration Model for Rt}.
\subsubsection{Spatiotemporal Dynamic of SGC}\label{Spatiotemporal Dynamic Model of SGC}
We first model the dendritic distribution of SGC neurons, taking an SGC-I neuron as an example. For a position at a distance $r$ from the center of the dendritic field $D$, the dendritic distribution density can be described as $\rho \left ( r \right )=\alpha \cos \left ( c\cdot r \right )+b$, $\alpha$, $b$ and $c$ are constants that can be determined by biological experiments \citep{ref8}. Assuming that an SGC parent node generates two symmetric dendrites and that the probability of occurrence of parent nodes is mutually independent, the probability of occurrence of a parent node in the dendritic field in region is $P_{c}\left ( A \right )=\frac{1}{2}\rho \left ( r \right )\cdot A$, where $A\subseteq D$. The positions of the terminals of two symmetrical dendrites follow a Gaussian distribution $g_{p}$. Therefore, for any position $x$ in the region $A$ of the dendritic field, the probability of the presence of dendrites is $P_{d}=g_{p}\cdot P_{c}\left ( A \right )$.

Building on the static distribution formulation above, we further derive a dynamic formulation for the special distribution of dendritic terminal spatial density. In the presence of external stimuli, signal transmission between RGC and SGC-I neurons can be described by the time-varying response probability, denoted $P_{r}\left ( \Delta t \right )$, where $\Delta t$ is the spike firing interval of the RGC. For a stationary stimulus, the probability of the number of activated dendritic terminals in the stimulated region $A$ being 
$n_{a}$is described as

\begin{eqnarray}\label{1}
P_{a}=\left ( \frac{\left ( \rho \left ( r \right )\cdot A \right )^{n_{a}}}{n_{a}!} \right )e^{-\rho \left ( r \right )\cdot A}
\end{eqnarray}

The neural response is given by $y_{s}=n_{a}\cdot P_{r}\left ( \Delta t \right )$ \citep{ref8}. For a moving stimulus, $y_{m}=y_{s}+P_{max}$, where $P_{max}=P\left ( \Delta t=\infty  \right )$ When the additional activated dendritic terminals are $n_{a}'$, the probability of at least one spike occurrence is $1-\left ( 1-P_{max} \right )^{n_{a}'}$, and the expected value of activations at this time step is
\begin{eqnarray}\label{2}
E_{a}=\sum_{{n_{a}'=1}}^{+\infty }{n_{a}'}\left [ P_{a} \cdot 1-\left ( 1-P_{max} \right )^{n_{a}'} \right ]
\end{eqnarray}

At time $t$, when a dendrite $s_{t}$ is activated, it generates energy $e_{t}$ and produces a corresponding energy radiation field $f_{t}$. For the next time step $t+1$, assuming the probability of dendrite $s_{t+1}$ falling into $f_{t}$ is $p_{t}$, and the energy generated when a dendrite is activated is a constant value $e'$, the energy $e_{t+1}=e_{t}+e'\cdot p_{t+1}$, not only includes the energy generated by its own activation but also accumulates the energy $e_{t}$ from the previous time step \citep{ref16}.

\subsubsection{Two-Stage Motion Information Integration of Rt}\label{Two-Stage Motion Information Integration Model for Rt}
% Figure
\begin{figure*}[htbp]
\centering
\label{fig1}
\subfigure[]
{
    \label{1.a}
    \begin{minipage}[b]{.26\linewidth}
        \flushleft\hspace{-0.8cm}
        \includegraphics[scale=0.51]{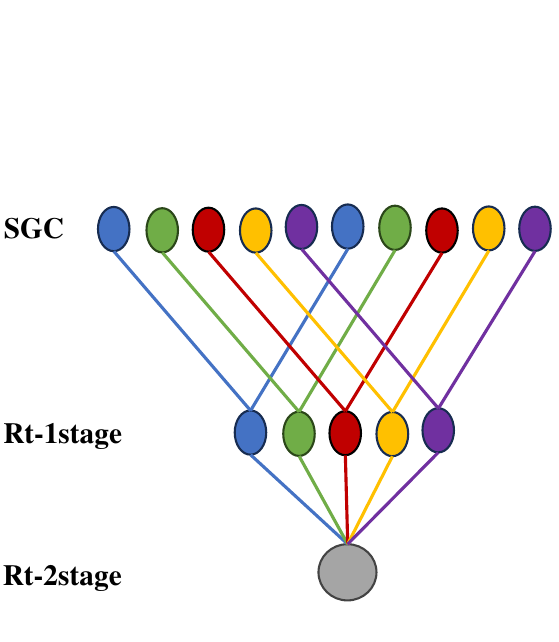}
    \end{minipage}
}
\subfigure[]
{
    \label{1.b}
    \begin{minipage}[b]{.3\linewidth}
        \flushleft\hspace{-0.6cm}
        \includegraphics[scale=0.48]{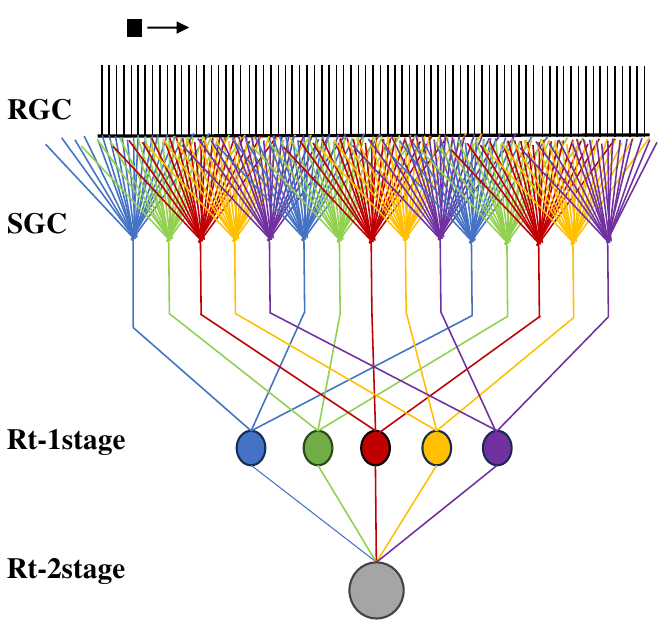}
    \end{minipage}
}
\subfigure[]
{
    \label{1.c}
 	\begin{minipage}[b]{.3\linewidth}
        \centering
        \includegraphics[scale=0.48]{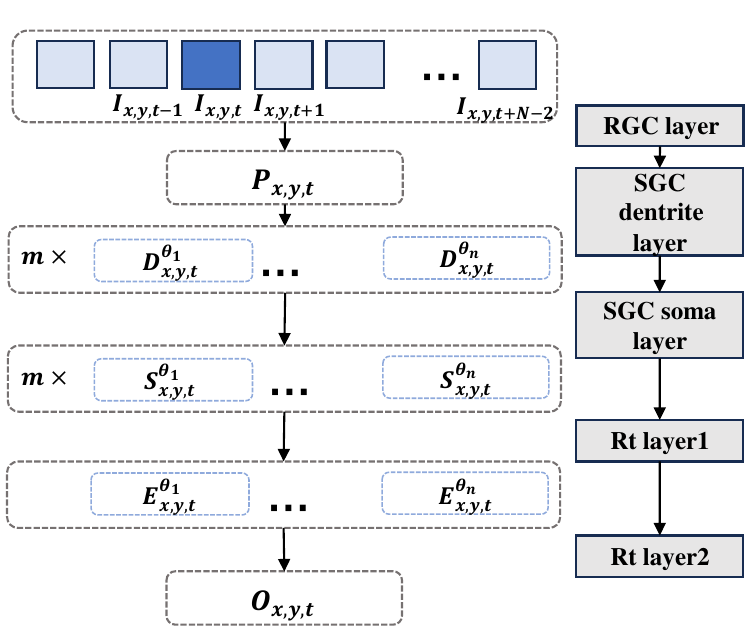}
    \end{minipage}
}	
\caption{Neural pathway and model architecture of RGC-SGC-Rt neural pathway. (a)SGC-Rt two-stage connection graph. (b)RGC-SGC-Rt neural connection graph. (c)RGC-SGC-Rt neural network model.}

\end{figure*}

Projection from SGC to Rt neurons cannot be described as a simple direct sampling; it requires introducing an intermediate point $M$, turning it into a two-stage process. As shown in the \cref{1.a}, intermediate points are sampled in $S$ at a fixed interval $\delta d$, with the number of intermediate points $N_{m}$ determined by $\delta d$, and $N_{m}<N_{s}$ (the number of SGCs). Each $x_{i}$ ($1\leq i\leq N_{s}$) is sampled by one intermediate point, forming a surjection, and the intermediate points further converge onto the Rt neuron sites. Divide $S$ into $N_{d}$ subsets, where each subset $S_{i}\left ( i\in \left [ 1,N_{d} \right ] \right )$ consists of elements spaced apart by $\Delta d$, then
\begin{eqnarray}\label{3}
\begin{aligned}
&N_{d}=\left \lfloor \frac{\left | S \right |}{\Delta d} \right \rfloor,\forall S_{i}\subseteq S\\
&S_{i}=\left \{ x_{1}, x_{1+\Delta d},...,x_{1+\left ( N_{d}-1 \right )\Delta d}\right \}
\end{aligned}
\end{eqnarray}

The expected activation of an output site $y_{j}$ in the one-stage framework is given by
\begin{eqnarray}\label{4}
E_{1}=\sum_{i=1}^{N_{d}}p_{i}e_{i}
\end{eqnarray}
where $p_{i}$ is the probability of selecting subset $S_{i}$, and $e_{i}$ is the probability that subset $S_{i}$ can activate $y_{j}$. For the two-stage framework, the expected activation of an output site $y_{j}$ is given by
\begin{eqnarray}\label{5}
E_{2}=1-\prod_{i=1}^{N_{d}}\left ( 1-e_{i} \right )
\end{eqnarray}

\newtheorem{thm}{Proposition}
\begin{thm}
For an output Rt neuron site, let $E_{2}$ be the activation probability of the two-stage framework and $E_{1}$ be that of the one-stage framework, then $E_{1}\leq E_{2}$.
\end{thm}
Proof. See Appendix \ref{A}.

\subsection{TSOM Neural Network Model for Small Object Motion Detection}\label{Retina-OT-Rt Neural Model for Small Object Motion Detection}

\begin{figure*}[tbp]
\centering
\includegraphics[scale=0.5]{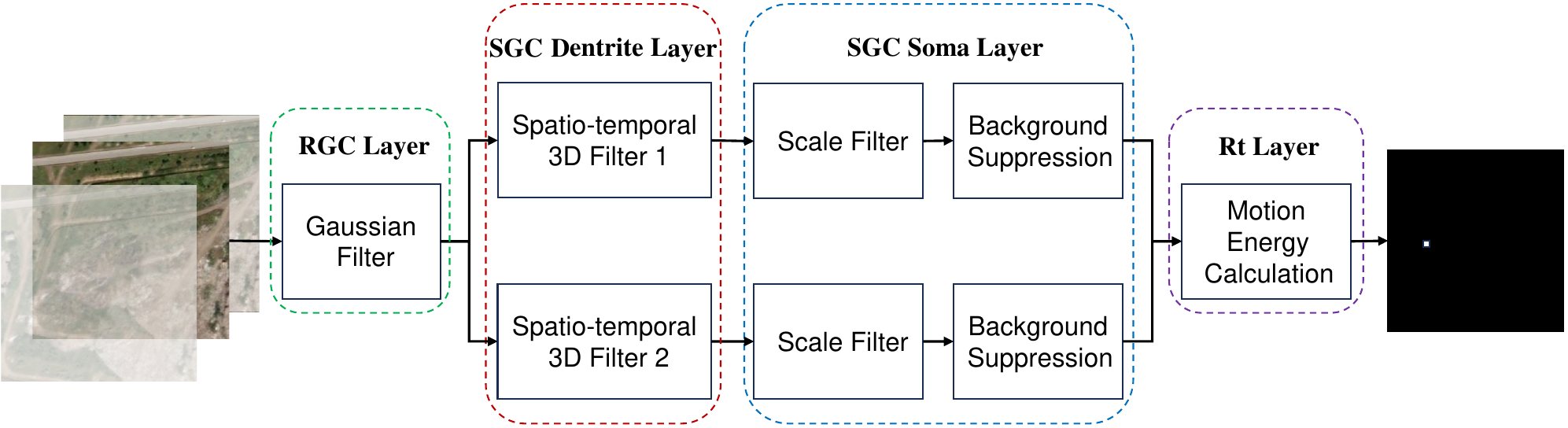}
\caption{Algorithm flow chart.}
\label{fig9}
\end{figure*}

Based on the mathematical description of the neural circuit, the proposed brain-inspired biological neural network model is composed of four subsystems, including the retina layer, SGC dendritic layer, SGC Soma layer, and two-stage Rt layer (including Rt layer1 and Rt layer2), as illustrated in Fig. \ref{1.c}. Once an image is received at time t, it is first received and pre-processed in the retinal layer, then applied to the SGC module to encode spatiotemporal information and compute motion information to extract small objects. In SGC layers, potential small objects are selected by a 3D spatiotemporal filter, and convolution with the spatial kernels and their contrast to the background is enhanced by the Z-score of the convolutional outputs. The contrast-enhanced image is fed into the Rt module for discriminating small moving objects from a complex background. The flowchart of the model is shown in Fig. \ref{fig9}. We introduce the neural model network components in Section \ref{Retina Layer}-\ref{Rt Layer}.
\subsubsection{Retina Layer}\label{Retina Layer}
The model proposed in this paper utilizes image sequences as input, necessitating the initial construction of a mapping from pixels to photoreceptors. As depicted in Fig. \ref{1.b}, each small square represents a pixel corresponding to a photoreceptor, and the entire image represents the full field of view. The yellow squares indicate individual pixels (photoreceptors). The response of RGC neurons to input stimuli can be approximated by a Gaussian smoothing of the input image \citep{ref18}. For an input image sequence with dimensions $H\times W$ and duration $T$, within the entire field of view $A$, the stimulus strength at any point is denoted as $I\left ( x,y,t \right )\in \mathbb{R}$, where $\left ( x,y \right )$ represents spatial coordinates, and $t$ denotes time. Given a Gaussian kernel with variance $\sigma _{1}^{2}$
\begin{eqnarray}\label{eq10}
G_{\sigma _{1}}\left ( x,y \right )=\frac{1}{2\pi \sigma _{1}^{2}}e^{\left ( -\frac{x^{2}+y^{2}}{2\sigma _{1}^{2}} \right )}
\end{eqnarray}

Then the output $P\left (x, y, t\right )$ of RGC neurons at point $\left ( x,y \right )$ can be defined as
\begin{eqnarray}\label{eq11}
P\left (x, y, t\right )=\iint I\left ( u,v,t \right )G_{\sigma _{1}}(x-u,y-v)dudv
\end{eqnarray}
\subsubsection{SGC Dendrite Layer}
To model the sensitivity of SGC neuron dendrites to changes, this section presents several sets of three-dimensional filters $F_{\varphi }$ with different preferred directions $\theta$ to detect variations in the input. For each direction $\theta$, there corresponds a 3D spatiotemporal filter $f_{\theta ,\varphi }\left ( x,y,t \right )$, with its spatial two-dimensional component being $f_{\theta ,\varphi }\left ( x,y \right )=g_{\theta ,\varphi }\left ( x,y \right )+h_{\theta ,\varphi }\left ( x,y \right )$, here $g_{\theta ,\varphi }\left ( x,y \right )$ and $h_{\theta ,\varphi }\left ( x,y \right )$ are mutually independent. 
\begin{eqnarray}\label{eq12}
\begin{aligned}
&g_{\theta ,\varphi }\left ( x,y \right )=e^{-\frac{x'\left ( \theta  \right )^{2}+\gamma ^{2}y'\left ( \theta  \right )^{2}}{2\sigma ^{2}}}e^{i\left ( 2\pi \frac{x'\left ( \theta  \right )}{\lambda } +\varphi \right )}\\
&x'\left ( \theta  \right )=x\cos \theta +y\sin \theta \\
&y'\left ( \theta  \right )=-x\sin \theta +y\cos \theta 
\end{aligned}
\end{eqnarray}
where $x'\left ( \theta  \right )$ and $y'\left ( \theta  \right )$ are the coordinates in the preferred direction $\theta$, and $\gamma$, $\sigma ^{2}$, $\lambda$ respectively represent the spatial aspect ratio, standard deviation, and wavelength of the 2D Gabor filter, and are constants.

A 2D spatial pattern moving at a particular velocity corresponds to a 3D spatiotemporal pattern in a given direction, which can be detected with a 3D spatiotemporal filter in the appropriate direction \citep{ref19}. Then
\begin{eqnarray}\label{eq15}
f_{\theta ,\varphi }\left ( x,y,t \right )=f_{\theta ,\varphi }\left ( x,y \right )\otimes i\left ( t \right )
\end{eqnarray}
where $i\left ( t \right )$ is the temporal kernel function for detecting changes in pixel values over time. The Sobel \citep{ref20} operator is employed in this paper.

For the input $P\left (x, y, t\right )$ to this layer, convolution is used to compute the mapping of each point along the preferred direction $\theta $
\begin{eqnarray}\label{eq16}
D_{\theta ,\varphi }\left ( x,y,t \right )=P\left (x, y, t\right )\ast f_{\theta ,\varphi }\left ( x,y,t \right )
\end{eqnarray}

\begin{figure}[bp]
\centering
\subfigure[]
{
 	\begin{minipage}[b]{.45\linewidth}
        \centering
        \includegraphics[scale=0.5]{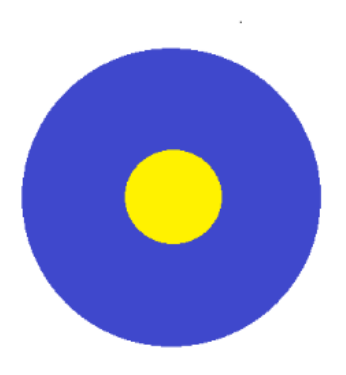}
    \end{minipage}
}
\subfigure[]
{
 	\begin{minipage}[b]{.45\linewidth}
        \centering
        \includegraphics[scale=0.5]{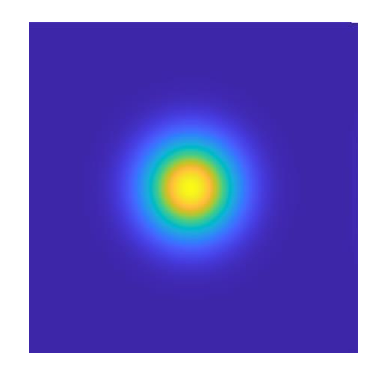}
    \end{minipage}
}	
\caption{Diagram of the scale selection kernel function. (a) Schematic representation of the receptive field of a neuron; (b) the scale selection kernel function.}
\label{fig2}
\end{figure}

\subsubsection{SGC Soma Layer}
$D_{\theta,\varphi }\left ( x,y,t \right )$ can detect the motion of objects at various scales and forms the input to this layer. According to the scale-selective characteristics of SGC neurons, scale filtering and background suppression are required for small object motion detection. To search for potential small objects, the spatial kernel function is defined as
\begin{eqnarray}\label{eq17}
W_{s}\left ( x,y \right )=e^{-a\left ( x^{2}+y^{2} \right )}-\mu e^{-\left ( x^{2}+y^{2} \right )}
\end{eqnarray}
where $\mu$ is a constant used to adjust the degree of suppression. As shown in \cref{fig2}, the spatial kernel measures the motion energy difference between the central part and surrounding areas. When the scale of object motion energy is smaller than the response area (yellow portion), the kernel function strongly responds. Moreover, the smaller the object, the greater the weight of the kernel function. However, the neuron response will be inhibited when it exceeds the response area (blue portion). Consider a spatial lateral inhibition scale-selective kernel function and use convolution computation for filtering
\begin{eqnarray}\label{eq18}
\begin{aligned}
S'_{\theta,\varphi }\left ( x,y,t \right )=\left [ {\iint D_{\theta,\varphi }\left ( x,y,t \right ) \left ( u,v,t \right )}\right.\\
\left.{\times W_{s}\left (x-u,y-v  \right )dudv}\right ]^{+}
\end{aligned}
\end{eqnarray}
where $S'_{\theta,\varphi }\left ( x,y,t \right )$ donates the feature map of motion energy for a certain direction. 

To enhance the contrast between potential small objects and the surrounding environment, we treat it as an anomaly detection problem. Taking the z-score calculation for observed points as an example, we present a background suppression method based on a statistical model. For each $S'_{\theta ,\varphi }$ with the time variable fixed, calculate the standard deviation of its distribution
\begin{eqnarray}\label{eq19}
\delta =\sqrt{\frac{1}{p\times q}\sum_{i=1}^{p}\sum_{i=1}^{q}\left ( S'_{\theta ,\varphi }-mean\left ( S'_{\theta ,\varphi }\right ) \right )^{2}}
\end{eqnarray}
where $mean\left ( \cdot  \right )$ denotes the mean. Then, for each observed point, its z-score is given by
\begin{eqnarray}\label{eq20}
Zscore_{\theta ,\varphi }\left ( x,y,t \right )=\frac{S'_{\theta ,\varphi }-mean\left ( S'_{\theta ,\varphi }\right )}{\delta }
\end{eqnarray}

Set a threshold $\epsilon $, where points below the threshold are considered the background, and points above the threshold are considered object points. The final output image after background suppression is 
\begin{eqnarray}\label{eq21}
S_{\theta ,\varphi }\left ( x,y,t \right )=S'_{\theta ,\varphi }\left ( x,y,t \right )\times \left [ Zscore_{\theta ,\varphi }\left ( x,y,t \right )-\epsilon  \right ]^{+}
\end{eqnarray}
\subsubsection{Rt Layer}\label{Rt Layer}
The projection from SGC to Rt neurons follows a two-stage model. In the initial phase, intermediate points combine inputs from subsets of SGC neurons that span the entire visual field. Subsequently, multiple intermediate points are projected onto an Rt neuron as a collective entity. The Rt neuron's output of motion features is generated through the synthesis of motion speed and direction. We simulate the integration function of intermediate points based on the motion energy model \citep{ref19}. The subset of SGC neurons that input to intermediate points is classified according to the motion direction $\theta$. For each $\theta$, the output feature maps, which exhibit a phase difference of $\frac{\pi }{2}$, are identified as an orthogonal pair in the motion energy model. The energy of motion for a specific point is defined as
\begin{eqnarray}\label{eq22}
E'_{\theta }\left ( x,y,t \right )=\sqrt{S'_{\theta ,\varphi }\left ( x,y,t \right )^{2}+S'_{\theta ,\varphi+\frac{\pi }{2} }\left ( x,y,t \right )^{2}}
\end{eqnarray}

Further, calculate the normalized net motion energy for each direction
\begin{eqnarray}\label{eq23}
E_{\theta }\left ( x,y,t \right )=\frac{E'_{\theta }\left ( x,y,t \right )}{E_{flk}}
\end{eqnarray}
\begin{eqnarray}\label{eq24}
E_{flk}=\frac{1}{\left | M \right |}\int_{M}^{}E_{\theta }\left ( x,y,t \right )
\end{eqnarray}
where $E_{flk}$ is referred to as flicker energy \citep{ref21}, defined as the average of the output energy, and $M$ is the number of feature maps.

The projection of neurons from SGC to Rt neurons causes an increase in its receptive field, which results in the precise spatial map becoming coarser in the Rt. To simulate the process of increased receptive field for $E_{\theta }\left ( x,y,t \right )$, the max-pooling method is used \citep{ref22}. Using this calculation, the model can retain more texture information of small objects and suppress smooth background information. Additionally, the output values of $E_{\theta }\left ( x,y,t \right )$ from various directions are combined, creating a simulation of the projection of all intermediate points in the Rt neurons. The customary information integration mechanism for neurons is a weighted sum, which produces the output image for the Rt as
\begin{eqnarray}\label{eq25}
O\left ( x,y,t \right )=\sum_{\theta }^{}\alpha _{\theta }E_{\theta }\left ( x,y,t \right )
\end{eqnarray}
where $\alpha _{\theta }$ represents the weights for different directions. $O\left ( x,y,t \right )$ is the ultimate motion feature map with motion speed and direction values (directions indicated by different $\theta$) at each point while maintaining spatial position information. This model is specifically designed to extract motion features associated with small objects, and therefore, the detected positions of small objects are indicated as
\begin{eqnarray}\label{eq26}
\left ( x_{s},y_{s},t\right )=argmax\left (O\left ( x,y,t \right )\right )
\end{eqnarray}
where $x_{s}$ and $y_{s}$ are the horizontal and vertical coordinates of the small moving object detected at time $t$.

\section{Experiment and Results}\label{Experiment and Results}
In this section, the TSOM model is experimentally validated from two perspectives: a) the biological consistency between TSOM layers and corresponding neurons is verified on synthetic image data constructed in this chapter; b) the model's performance on detecting moving small objects in different scenes composed of multiple factors is discussed on both synthetic and natural image datasets, compared with three other unsupervised moving object detection methods to verify that TSOM model has higher accuracy for small object detection on image sequences. The simulation experiments are conducted on an NVIDIA RTX 3090 GPU.

\subsection{Datasets and Metrics}\label{Datasets}
\subsubsection{Neuroelectrophysiological data}
Utilizing pigeons as a model organism, our study delves into the response characteristics of avian OT neurons to the motion features of small objects. Through a meticulously designed experimental paradigm, we acquired neural response data of OT neurons under diverse scale object motion stimuli and, correspondingly, OT neuron response data under varying speeds of small objects. Drawing upon neurophysiological data, we further deduced the response patterns of OT neurons to object scale and object speed. (Experimental paradigm and experimental setup refer to Appendix \ref{app:Response properties of small target motion in pigeon OT neurons}.) The aforementioned experimental data and response patterns will be employed in this study to validate the biological consistency of TSOM.
\subsubsection{Image sequence data}
\begin{figure*}[htbp]
\centering
\subfigure[]
{
        \label{4.a}
 	\begin{minipage}[b]{1\linewidth}
        \centering
        \includegraphics[scale=0.7]{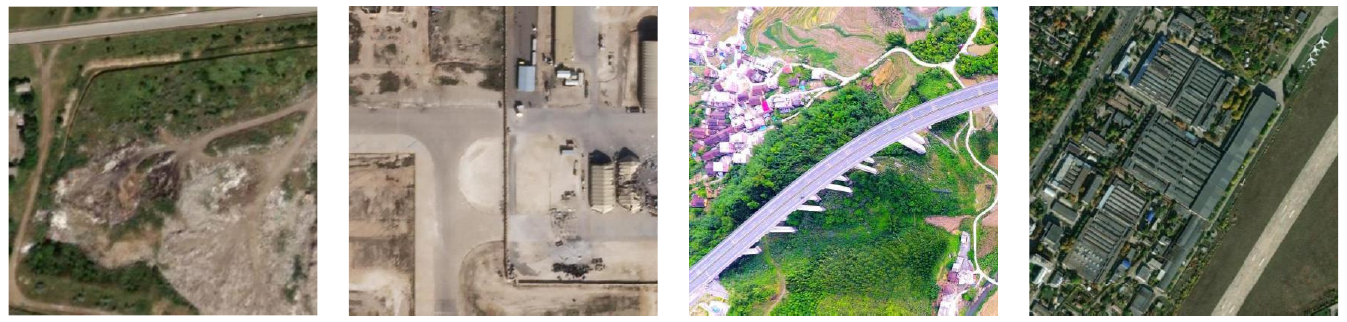}
    \end{minipage}
}
\subfigure[]
{
        \label{4.b}
 	\begin{minipage}[b]{1\linewidth}
        \centering
        \includegraphics[scale=0.7]{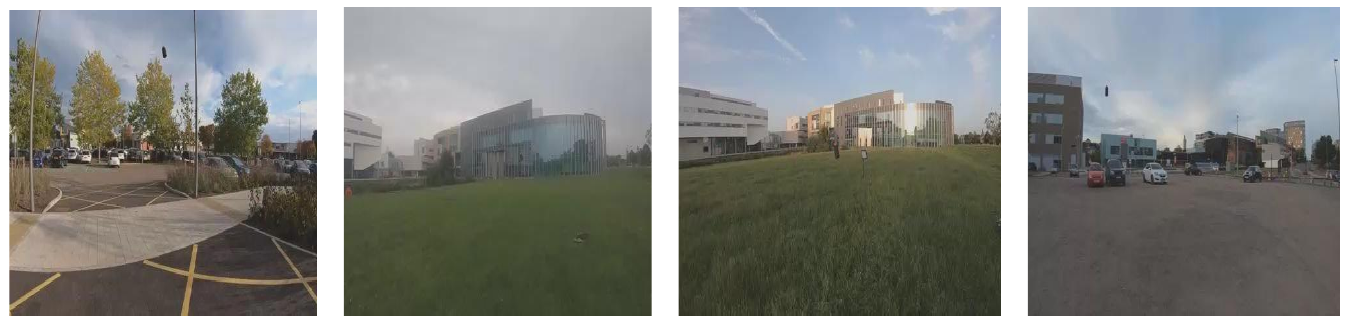}
    \end{minipage}
}	
\caption{Examples from the datasets of (a) BEV and (b) RIST.}
\label{fig3}
\end{figure*}
We utilize a synthetic dataset BEVS (Bird Eye View with Synthetic Object) and a real-world dataset known as RIST \citep{ref23} to assess the effectiveness of our proposed model, TSOM, in detecting small object motion.

Considering the unique scenarios the avian visual system encounters, we generate a synthetic dataset of small objects using satellite imagery. The dataset comprises satellite images as backgrounds, with each frame measuring 512×512 pixels. Within each frame is a solitary object with a radius of 3 pixels, occupying merely $0.003\%$ of the image size. The dataset encompasses 12 image sequences that simulate birds observing their surroundings during mid-flight. In these sequences, the background moves at a consistent speed and direction, presenting challenges such as extremely minuscule and low-contrast objects amidst highly intricate dynamic backgrounds. \cref{4.a} illustrates several sample scene images from the BEVS dataset.

The RIST dataset has been developed as a real-world dataset specifically designed to detect small moving objects. It comprises 19 video sequences captured using a motion camera, with each frame having dimensions of 480×270 pixels and featuring a solitary object. The size of the objects varies from 3×3 to 5×5 pixels. These video scenes exhibit intricate moving backgrounds, subtle objects, diverse lighting, weather conditions, and sudden environmental changes. \cref{4.b} showcases several sample scene images for reference purposes.
\subsubsection{Metrics}
The model's biological validation experiments were conducted during the subsequent experimental sessions using the pixel points' values in the current feature map as output values. In the small object extraction performance experiments, the prediction order was determined by the order of the pixel point values in a map. A predicted point within 5 pixels of the actual point was considered a positive example. 

In the subsequent experimental sessions, the biological validation experiments of the model are conducted using the output values of the pixel points in the current feature map. To quantitatively evaluate performances of small object motion detection methods, we adopt two comparison metrics, i.e.,  detection rate ($D_{R}$) and false alarm rate ($F_{A}$) \citep{ref31,ref33,ref34,ref28}. If the distance of a detected position to a ground truth is within a threshold (5 pixels), we declare it as a true positive. Then, the two metrics can be formulated as
\begin{eqnarray}\label{eq27}
\begin{aligned}
&D_{R}=\frac{number\:of\:true\:positives}{number\:of\:actual\:objects}\\
&F_{A}=\frac{number\:of\:false\:positives}{number\:of\:frames}
\end{aligned}
\end{eqnarray}
where $D_{R}$ measures the percentage of small objects that can
be detected correctly, while $F_{A}$ represents the mean number of false positives in each frame. The receiver operating characteristics (ROC) curves are computed in each experiment.

\subsection{Bio-consistency validation}
This section verifies the biological consistency between each layer of the TSOM model and the neuronal responses in two ways. Firstly, validation experiments for signal processing of the TSOM model pathway are conducted to confirm the consistency between the model layers and the neuronal response in the visual pathway when exposed to external stimuli. Secondly, validation experiments for model response properties were conducted to confirm that the selectivity of the relevant model layers for scale and velocity is consistent with the results of related neuroscience experiments.
\subsubsection{Effectiveness of Signal Proceccing}
\begin{figure*}[tbp]
\centering
\subfigure[]
{
        \label{5.a}
 	\begin{minipage}[b]{.45\linewidth}
        \flushright
        \includegraphics[scale=0.4]{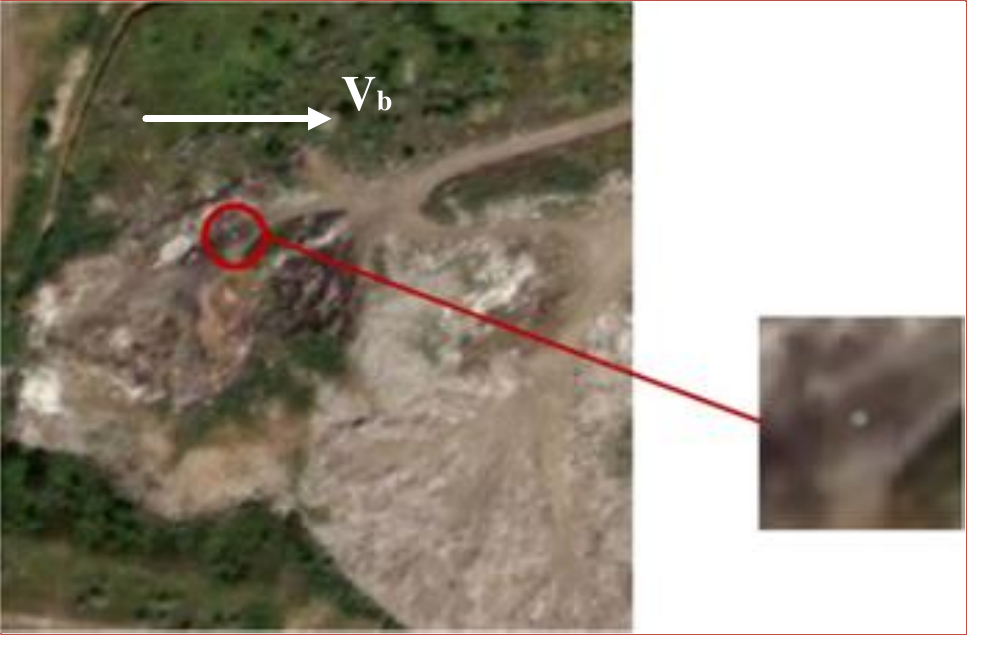}
    \end{minipage}
}
\subfigure[]
{
\label{5.b}
 	\begin{minipage}[b]{.45\linewidth}
        \centering
        \includegraphics[scale=0.4]{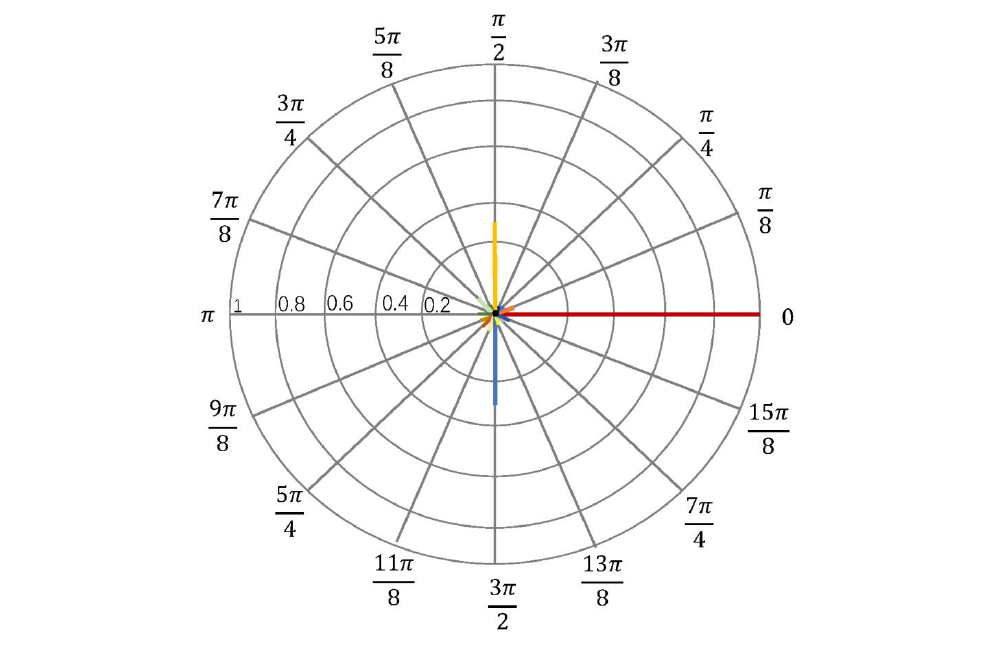}
    \end{minipage}
}
\subfigure[]
{
\label{5.c}
 	\begin{minipage}[tb]{.45\linewidth}
        \centering
        \includegraphics[width=7.5cm,height=2.5cm]{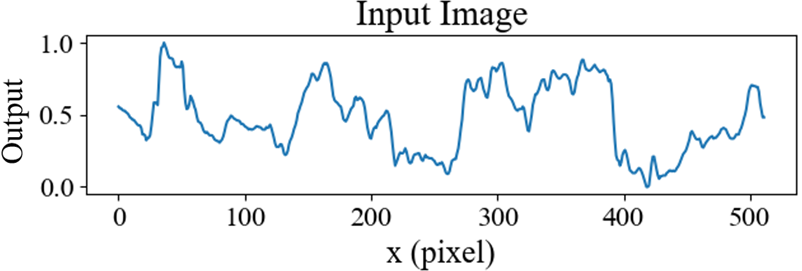}
    \end{minipage}
}	
\subfigure[]
{
\label{5.d}
 	\begin{minipage}[tb]{.45\linewidth}
        \centering
        \includegraphics[width=7.5cm,height=2.5cm]{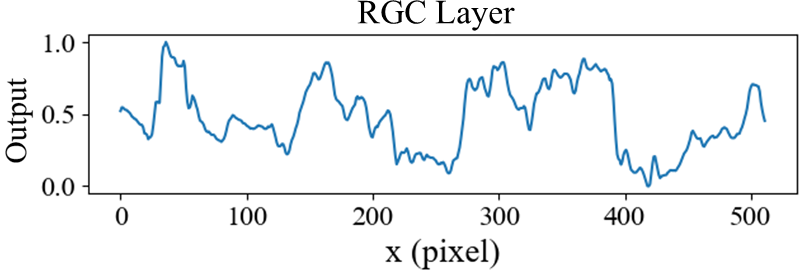}
    \end{minipage}
}
\subfigure[]
{
\label{5.e}
 	\begin{minipage}[tb]{.451\linewidth}
        \centering
        \includegraphics[width=7.5cm,height=2.5cm]{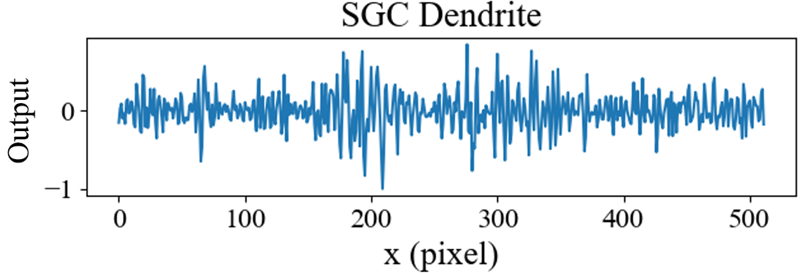}
    \end{minipage}
}
\subfigure[]
{
\label{5.f}
 	\begin{minipage}[tb]{.45\linewidth}
        \centering
        \includegraphics[width=7.5cm,height=2.5cm]{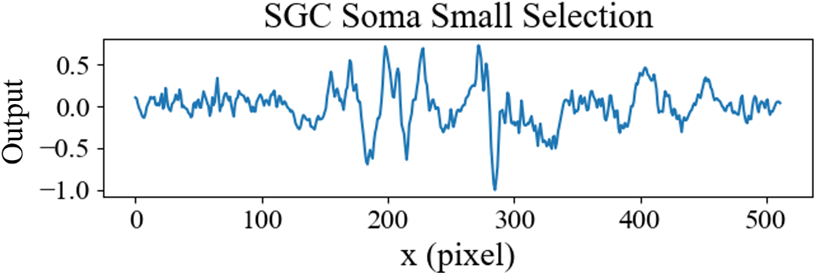}
    \end{minipage}
}

\subfigure[]
{
\label{5.g}
 	\begin{minipage}[tb]{.45\linewidth}
        \centering
        \includegraphics[width=7.5cm,height=2.5cm]{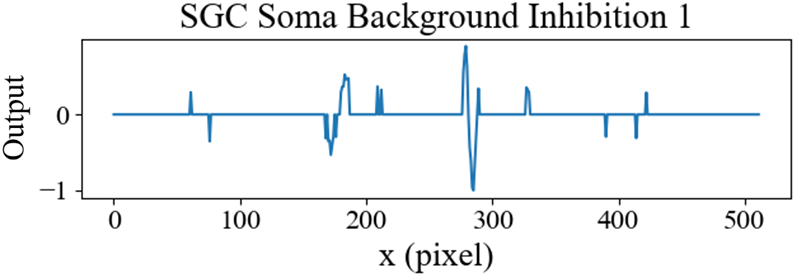}
    \end{minipage}
}
\subfigure[]
{
\label{5.h}
 	\begin{minipage}[tb]{.45\linewidth}
        \centering
        \includegraphics[width=7.5cm,height=2.5cm]{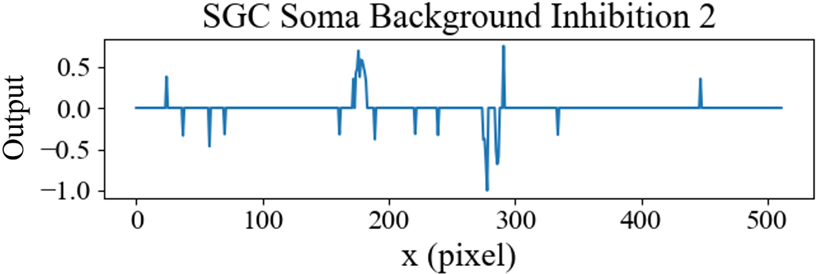}
    \end{minipage}
}
\subfigure[]
{
\label{5.i}
 	\begin{minipage}[tb]{.45\linewidth}
        \centering
        \includegraphics[width=7.5cm,height=2.5cm]{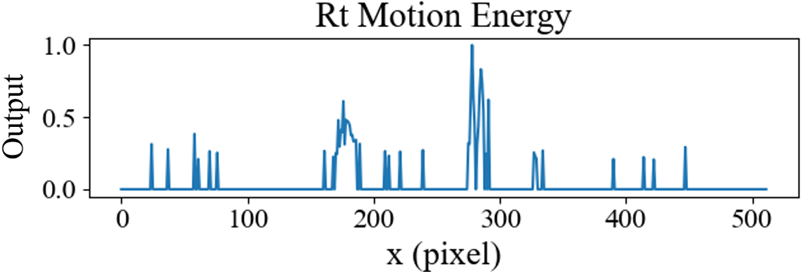}
    \end{minipage}
}
\subfigure[]
{
\label{5.j}
 	\begin{minipage}[tb]{.45\linewidth}
        \centering
        \includegraphics[width=7.5cm,height=2.5cm]{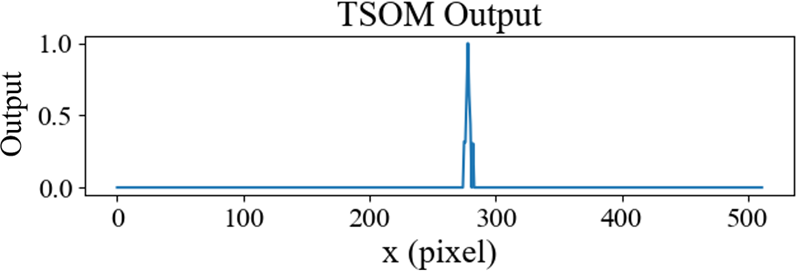}
    \end{minipage}
}
\caption{Signal processing validation experimental paradigm and results. (a) Input image $I\left ( u,v,t \right )$ with a resolution of 512×512. The background moves at a velocity of $V_{b}$ along a fixed direction, while the object moves at a velocity of $V_{a}$ along the direction $\theta=0$. The object's initial position is located at coordinates $\left ( 282,102 \right )$. (b) Results of directional selectivity validation. (c)-(j) The output of each module of the TSOM.}
\label{fig4}
\end{figure*}

To clearly illustrate the signal processing characteristics of each layer in the model pathway, this experiment examines the output of each neural layer to validate its efficacy. \cref{5.a} shows the input image $I\left ( u,v,t \right )$, corresponding to Scene 1 in the BEVS dataset consisting of 20 input frames with a resolution of 512×512. The background exhibits a constant motion at a speed denoted as $V_{b}$ along a fixed direction, while the object moves at a velocity of $V_{a}$ along the direction $\theta=0$. The object's initial position is located at coordinates $\left ( 282,102 \right )$. In \cref{5.c,5.d,5.e,5.f,5.g,5.h,5.i,5.j}, we present the output of each layer within TSOM concerning variations in $x$, keeping $y_{0}=102$ and $t_{0}=2$ fixed. To facilitate observation, all output values have been normalized for clarity.

The results in \cref{5.d} demonstrate that the output of the RGC layer undergoes Gaussian smoothing of the input signal brightness, aligning with the intended function of the layer. Furthermore, it is evident from \cref{5.e} that the output response of the SGC dendrite layer exhibits significant variations, indicating its high sensitivity to the spatiotemporal changes in pixels and capability to integrate dynamic information, which is consistent with the functional characteristics of the SGC dendrites. From the small object selection map, it can be observed that the application of a spatial lateral inhibition scale-selection kernel function, resembling the one illustrated in \cref{fig2}, results in a smoother output shown in \cref{5.f}. This effectively mitigates the impact of large-scale background motion. The specific impact of the background suppression module is detailed in \cref{5.g,5.h}, and it can be seen that post background suppression, the output value at the location of the small object $\left( x=282 \right)$ becomes significantly pronounced. Following the initial assumption that pixels in the background share similar directional and size motion characteristics, the background suppression operation further suppresses pixels conforming to the statistical distribution on the motion feature map, highlighting pixel points of small objects with motion characteristics different from the background.

As shown in \cref{5.i}, the numerical output of the Rt layer represents the motion energy in the direction of $\theta=0$. In this model, the Rt layer calculates motion energy through intermediate points, obtaining the motion energy map in the preferred direction $\theta$. In \cref{5.b}, it can be observed that for small objects with a motion direction $\theta=0$, the output of motion energy is significantly more potent at $\theta=0$ than in other directions, demonstrating strong direction selectivity. This indicates that the model effectively extracts motion features, and the output aligns with the direction selectivity characteristics of the Rt neurons. The final output, as shown in \cref{5.j}, indicates that the maximum response corresponds to the position of the small object, as can be seen, $x=282$. The processing of various layers in this model indicates that the TSOM model responds well to small objects and exhibits no response to larger objects.

\subsubsection{Response Properties of TSOM}
\begin{figure*}[htbp]
\centering
\subfigure[]
{
\label{6.a}
 	\begin{minipage}[tb]{.53\linewidth}
        \centering
        \includegraphics[scale=0.75]{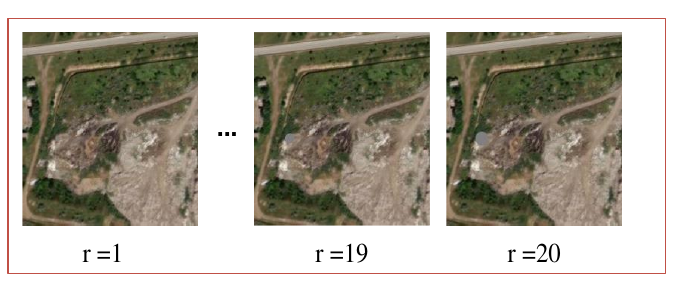}
    \end{minipage}
}
\subfigure[]
{
\label{6.b}
 	\begin{minipage}[tb]{.35\linewidth}
        \centering
        \includegraphics[scale=0.75]{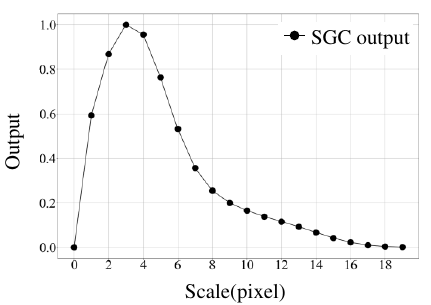}
    \end{minipage}
}	
\subfigure[]
{
\label{6.c}
 	\begin{minipage}[tb]{1\linewidth}
        \centering
        \includegraphics[scale=0.8]{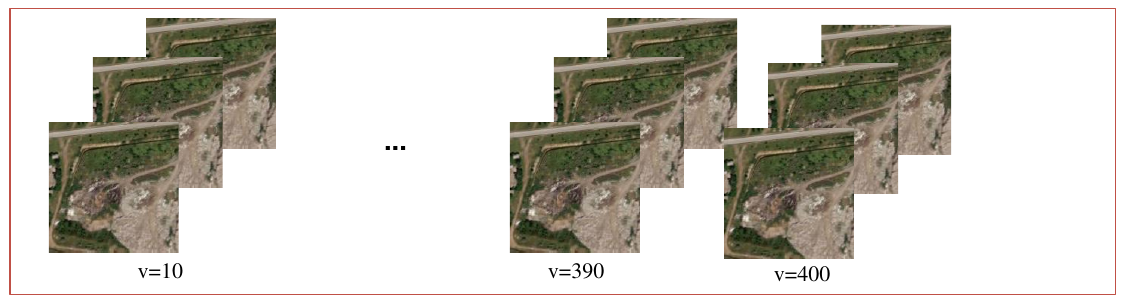}
    \end{minipage}
}
% \subfigure[]
% {
% \label{6.d}
%  	\begin{minipage}[tb]{.23\linewidth}
%         \centering
%         \includegraphics[scale=0.54]{fig5d.pdf}
%     \end{minipage}
% }
\subfigure[]
{
\label{6.e}
 	\begin{minipage}[tb]{.43\linewidth}
        \centering
        \includegraphics[scale=0.74]{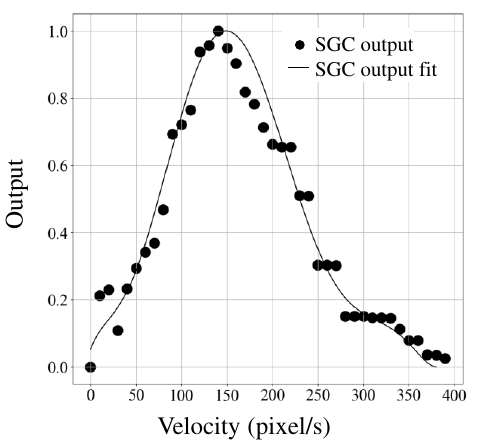}
    \end{minipage}
}
% \subfigure[]
% {
% \label{6.f}
%  	\begin{minipage}[tb]{.23\linewidth}
%         \centering
%         \includegraphics[scale=0.54]{fig5f.pdf}
%     \end{minipage}
% }
\subfigure[]
{
\label{6.g}
 	\begin{minipage}[tb]{.43\linewidth}
        \centering
        \includegraphics[scale=0.74]{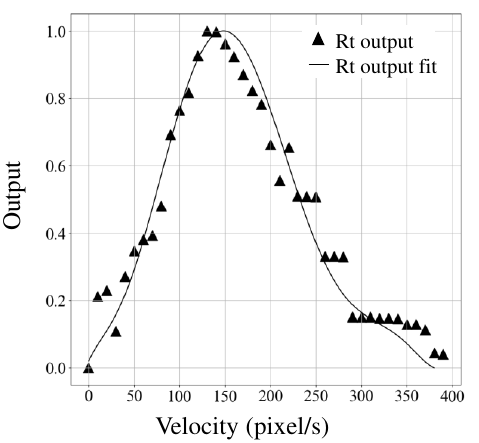}
    \end{minipage}
}
% \caption{Experimental paradigm and recorded outputs of the response properties simulation. (a) Object with different size; (b) SGC scale selectivity; (c) Object with different size; (d) SGC speed selectivity; (e) SGC velocity selective fitting; (f) Rt speed selectivity; (g) Rt velocity selective fitting.}
\caption{Experimental paradigm and recorded outputs of the response properties simulation. (a) Object with different sizes; (b) SGC scale selectivity; (c) Object with different sizes; (d) SGC velocity selectivity; (e) Rt velocity selectivity.}
\label{fig5}
\end{figure*}

To reveal the tuning properties of the TSOM neural
network, we report its outputs $E_{\theta }\left ( x,y,t \right )$ to a moving small object with different sizes and velocities. In the experiment concerning object size, the output of the SGC Soma layer was calculated with fixed background motion direction and velocity and an object radius ranging from 0 to 20. In the experiment on velocity, the background moved horizontally from left to right at a velocity of $V_{b}=150$ pixel/s, with the object moving at a velocity ranging from 10 to 400 pixel/s. The object radius equates to 3 pixels. The experimental paradigm and recorded outputs are shown in \cref{fig5}.

As observed from \cref{6.b}, the outputs of the SGC soma layer to objects with different radii are presented. It can be seen that there is a response to objects with a radius lower than 17. In addition, its output peaks at radius = 3 pixels. In \cref{6.e}, we can find that the output of the SGC soma layer is much larger than 0 in the interval $\left [ 0,400 \right ]$ pixel/s and reaches its maximum at 140 pixel/s. In \cref{6.g}, it is found that the output of the Rt layer is quite similar to the SGC layer, which corresponds to the preferred velocity range and optimal velocity of the TSOM model, respectively.

The biological experiments in Appendix \ref{Response Characteristics of OT Neurons to Small Object Motion} reveal that SGC neurons exhibit selectivity towards scale and velocity. Specifically, SGC neurons exhibit heightened responses towards neurons of more minor scales, as referenced in Appendix \ref{Small-Scale Targets}. Furthermore, they demonstrate increased responses towards swifter objects, as Appendix \ref{Velocity Preferences for Small Targets} indicates.

\cref{fig5} provides an excellent fit to the response properties of the SGC and Rt neurons revealed in biological research \citep{ref16,ref8,ref9,ref17} and Appendix \ref{Response Characteristics of OT Neurons to Small Object Motion}, which means the proposed TSOM model displays velocity and size selectivities, respectively.

\subsection{Small Object Detection Performance and Comparative Experiments}
This section presents experiments focusing on two aspects of the TSOM model's ability to extract small object positions. Firstly, the factors affecting the localization ability of the TSOM are discussed, including object scales, velocities, luminance, background velocities, and motion directions. Additionally, we compare the performance of the TSOM with three unsupervised motion object detection methods in both the BEV and RIST datasets.

\subsubsection{Parameter Sensitivity of the TSOM}
\begin{figure}[htbp]
\centering
\subfigure[]
{
\label{7.a}
 	\begin{minipage}[b]{.45\linewidth}
        \flushleft\hspace{-5mm}
        \includegraphics[width=4.2cm,height=3.5cm]{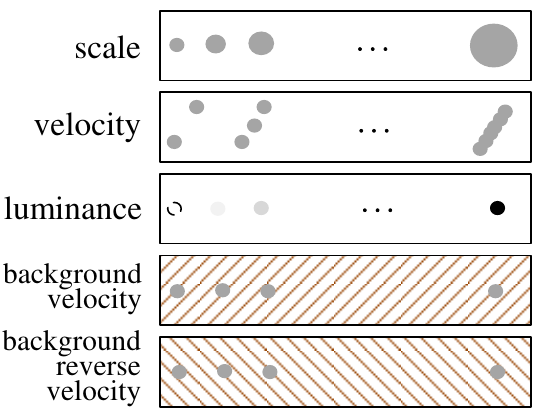}
    \end{minipage}
}
\subfigure[]
{
\label{7.b}
 	\begin{minipage}[b]{.45\linewidth}
        \centering
        \includegraphics[scale=0.45]{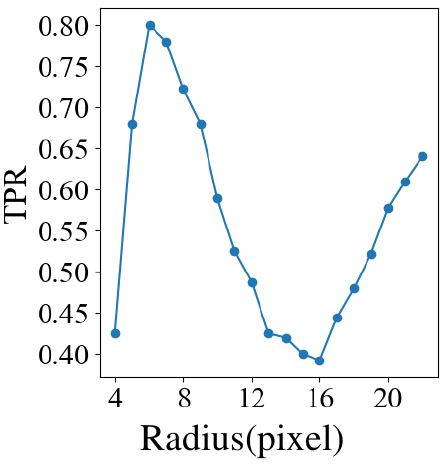}
    \end{minipage}
}	
\subfigure[]
{\label{7.c}
 	\begin{minipage}[tb]{.45\linewidth}
        \centering
        \includegraphics[scale=0.45]{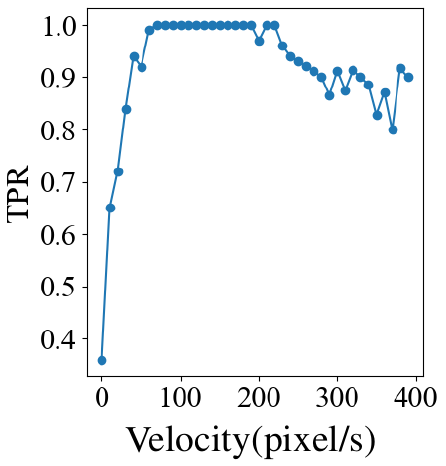}
    \end{minipage}
}
\subfigure[]
{\label{7.d}
 	\begin{minipage}[tb]{.45\linewidth}
        \centering
        \includegraphics[scale=0.45]{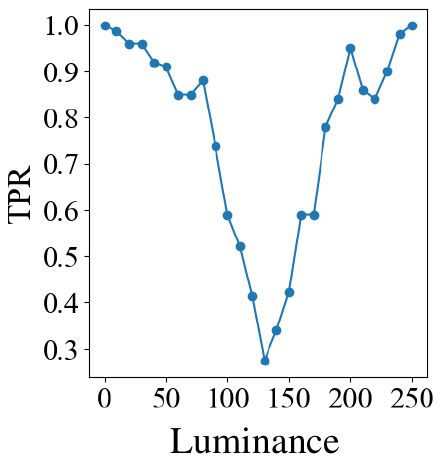}
    \end{minipage}
}
\subfigure[]
{\label{7.e}
 	\begin{minipage}[tb]{.45\linewidth}
        \centering
        \includegraphics[scale=0.45]{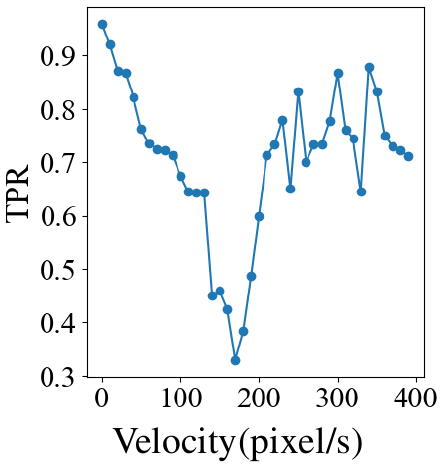}
    \end{minipage}
}
\subfigure[]
{\label{7.f}
 	\begin{minipage}[tb]{.45\linewidth}
        \centering
        \includegraphics[scale=0.45]{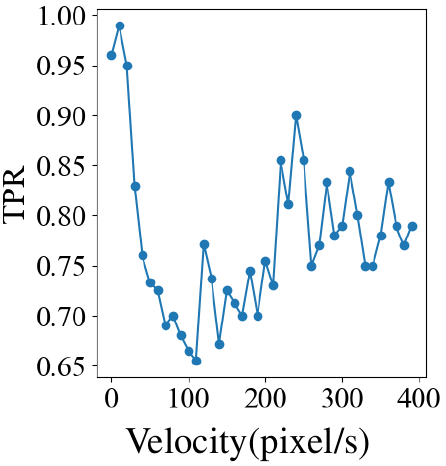}
    \end{minipage}
}
\caption{Experimental paradigm (a) and outputs of TSOM to a moving object concerning different (b) object scales, (c) object velocity, (d) object luminance, (e) background velocity, and (f) background direction for exploring factors that influence model performance.}
\label{fig6}
\end{figure}

Many factors can influence the accuracy of the model output, as shown in the \cref{tbl1} and \cref{7.a}. Based on five parameters: object radius, object motion velocity, object brightness, background motion velocity, and background motion direction, five sets of image sequences are generated using image sequence 1 from the BEVS dataset. Each trial consists of a continuous video of 200 frames. The distance $d$ between the detected and ground truth positions for each frame is calculated. Detection is considered correct if $d\leq5$. The average accuracy rate over the 200 frames is computed as each trial's evaluation metric. The model's performance across these five sets of image sequences is assessed, and Precision-Parameter curves are plotted, as shown in \cref{7.b,7.c,7.d,7.e,7.f}.

\begin{table}[htbp]
\centering\small
\caption{Parameters for synthesizing image sequences}\label{tbl1}
\begin{tabular}[b]{lccccc}
\toprule
\makebox[0.09\textwidth][l]{Parameters} & \makebox[0.05\textwidth][c]{Seq.1} & \makebox[0.05\textwidth][c]{Seq.2} & \makebox[0.05\textwidth][c]{Seq.3} & \makebox[0.05\textwidth][c]{Seq.4} & \makebox[0.05\textwidth][c]{Seq.5}\\
\midrule
$V_{a} $  &0-400   &150  &150  &150 &150 \\
$Radius $  &3  &1-20  &3   &3  &3\\
$Lm$  &0 &0 &0 &0 &0\\
$V_{b}$  &150 &150  &150 &0-400  &150 \\
$\theta$  &1 &1  &1 &1 &2 \\
\bottomrule
\end{tabular}
\end{table}
\normalsize

As shown in \cref{7.b}, when the object radius is in the range of 1 to 4 pixels, the accuracy rate rapidly increases with the enlargement of the radius. This can be attributed to the scale-selective characteristics of the SGC soma layer, as depicted in \cref{6.b}. The output of the SGC soma layer gradually increases to its maximum within the 1 to 4 scale range, making the detection and localization of objects more accessible, resulting in the highest correctness rate. Moreover, as can be seen, when the object radius exceeds the optimal selection value of the SGC soma layer, the output of the SGC soma layer gradually weakens, corresponding to a gradual decrease in the accuracy rate of small object detection. In the TSOM model, representing a single SGC neuron, the convolution kernel size is 13 pixels, which is selective for objects smaller than 5 pixels. Therefore, when the object radius is larger than 13 pixels, it exceeds the receptive field size of a single SGC neuron, activating surrounding neurons and consequently leading to a gradual increase in the correctness rate.

The model's response properties for object motion velocity are illustrated in \cref{7.c}. As the object motion increases, the accuracy shows an overall increasing trend. In the velocity range of 0 to 100 pixel/s, the correctness rate rapidly increases to 1 and maintains this high correctness rate within a specific velocity range, corresponding to the region in \cref{6.g} where the Rt output is larger than 0.8. However, due to the velocity selectivity caused by the refresh rate, the accuracy fluctuates as the object velocity continues to increase but maintains a relatively high value overall.

In the experiment investigating the influence of brightness factors, the brightness value of the background surrounding the object is set to 131. As shown in \cref{7.d}, when the object brightness is 0, it stands out in the background, resulting in an accuracy of 1. As the brightness gradually increases, making the object brightness closer to the background, the difficulty of detecting small objects increases, leading to a gradual decrease in accuracy. When the brightness increases, creating a higher contrast with the background, the small object becomes more prominent again, causing the accuracy to rise to 1 gradually.

As shown in \cref{7.e,7.f}, when the background motion velocity varies, the accuracy fluctuates significantly. When the background motion velocity is 0, the object's motion is more pronounced against a static background, making it more easily extracted. Hence, the accuracy is higher. When the background moves in the same direction as the object, as the background velocity gradually approaches the object's speed, the relative motion decreases, making it more difficult to detect the small object, resulting in a gradual decrease in the accuracy rate to a minimum of 0.33. As the background speed further increases, creating a greater difference in motion with the object, the background is suppressed by the background inhibition module of the SGC soma layer, leading to an increase in the accuracy, ultimately fluctuating around 0.75. When the background and object move in opposite directions due to different motion directions, there will be decoding on feature maps representing different directions in the Rt layer. Compared to co-directional motion, the velocity difference between the background and the small object moving in the opposite direction is larger. Therefore, the accuracy will be slightly higher than co-directional motion, fluctuating around 0.8.

\subsubsection{Evaluation on Synthetic and Real-World Datasets}
\begin{figure*}[tbp]
\centering
\includegraphics[scale=0.38]{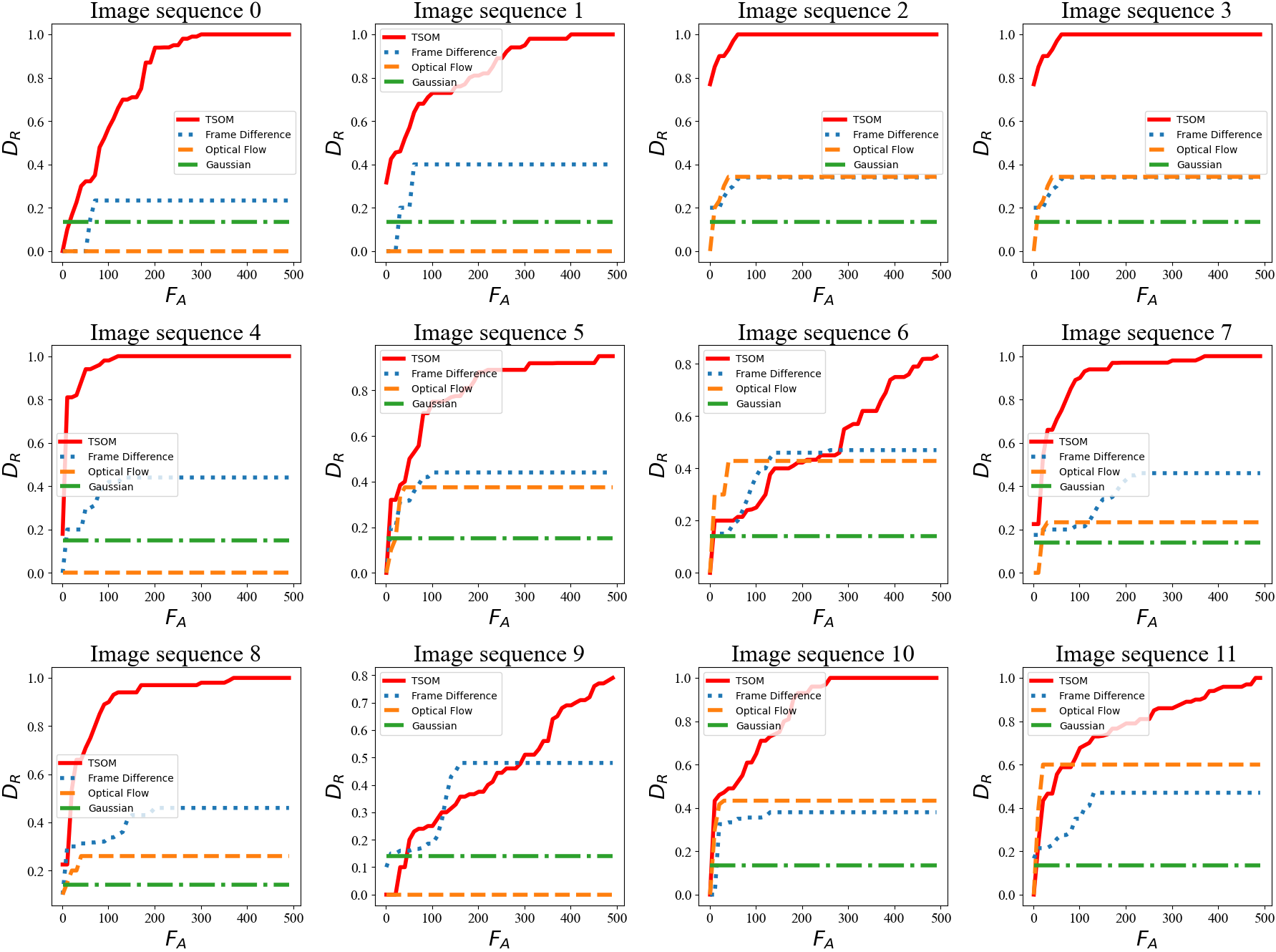}
\caption{Experimental results of multi-model comparison in BEVS dataset.}
\label{fig7}
\end{figure*}

\begin{table}[tbp]
\centering\small
\caption{Results of the models on BEVS dataset}\label{tbl2}
\begin{tabular}[b]{ccccc}
\toprule
\makebox[0.02\textwidth][c]{$F_{A}$} & \makebox[0.14\textwidth][c]{Frame Difference} & \makebox[0.07\textwidth][c]{Optical Flow} & \makebox[0.07\textwidth][c]{Gaussian} & \makebox[0.03\textwidth][c]{TSOM}\\
% $ $  &Frame Difference  &Optical Flow  &Gaussian  &TSOM  \\
\midrule
300  &0.4094   &0.2513  &0.1383  &\pmb{0.8941}  \\
\bottomrule
\end{tabular}
\end{table}

\begin{figure*}[tbp]
\centering
\includegraphics[scale=0.95]{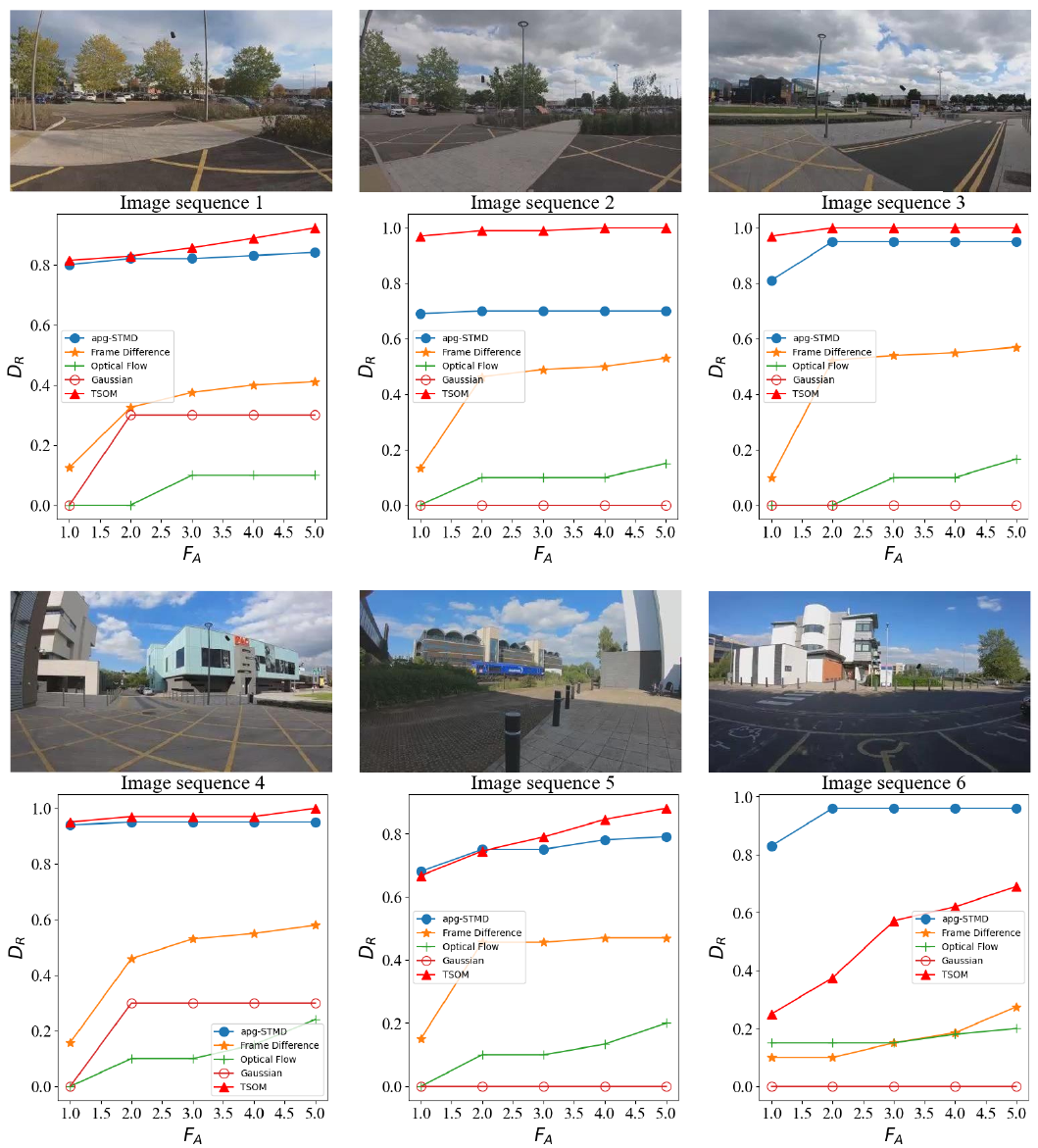}
\caption{Experimental results of multi-model comparison in RIST dataset.}
\label{fig8}
\end{figure*}

To assess the generalization performance of the proposed model in various scenarios, experiments were conducted on the synthetic dataset BEVS and the natural image dataset RIST. The model's performance was compared with classical computer vision methods, including Frame Difference Method \citep{ref25}, Optical Flow \citep{ref26}, and Mixture of Gaussians Background Modeling \citep{ref27}, on the synthetic dataset. In addition to the three algorithms, we also included the apg-STMD \citep{ref28}, a small object model detection based on the insect visual pathway, which is currently the most effective model in the STMD series. The evaluation was conducted on the real-world dataset RIST. The comparative results of the three benchmark algorithms and the proposed method on the BEVS dataset are presented in \cref{tbl2} and \cref{fig7}. The instance images from the RIST and experimental results are shown in \cref{fig8}.
% The apg-STMD algorithm is not listed because it does not work well on BEVS with the parameter settings recommended in their original paper. 

From \cref{fig7}, it can be observed that, for moving small objects in complex scenes at the same recall rate, the proposed method TSOM consistently achieves higher accuracy in small object detection across all scenarios in the dataset compared to the other three algorithms. The proposed method exhibits high accuracy and generalization performance for small object detection in high-altitude overhead scenes. From the \cref{fig8}, it is evident that the Frame Difference Method, Optical Flow, and Mixture of Gaussian background Modeling still perform relatively poorly on the RIST dataset. Although the apg-STMD algorithm has demonstrated its effectiveness, as reflected in all image sequences in \cref{fig8}, the proposed TSOM method performs better, particularly in image sequences 2, 3, and 4, where the accuracy is already close to 1 at a recall rate of 1. However, the proposed method's ability to extract small objects is affected when there is low contrast between the small object and the surrounding background, particularly in the presence of high-contrast areas in the background, leading to decreased accuracy.

\section{Conclusion}\label{Conclusion}
This article models the Retina-OT-Rt neural circuit for extracting motion features of small objects in the avian visual system. Building upon previous anatomical and neurophysiological studies on this circuit, we utilize mathematical derivations to analyze the response mechanisms of neurons in each layer of the circuit to the motion of small objects. This circuit's principles of small object extraction are elucidated from the perspectives of neuron distribution and connectivity specificity. Subsequently, a motion detection neural network model TSOM is constructed based on the response mechanisms, capable of effectively extracting motion features of small objects in complex scenarios observed from a high aerial view. The equivalency between the proposed model and the response mechanisms is analyzed, demonstrating the biological interpretability of the model. Through signal processing validation experiments and response characteristic simulations of small object extraction at each layer of the model, it is demonstrated that the TSOM model adheres to the tuning characteristics of neurons in the avian Retina-OT-Rt neural circuit, aligning with the biological functions of each layer's neurons. Finally, we compare the TSOM model with other models in extracting motion features of small objects using the synthesized BEVS dataset and the publicly available RIST dataset. The proposed model exhibits higher accuracy in small object extraction across various scenarios, validating its superior detection performance in diverse environments.

Although TSOM has shown promising performance on most datasets, a noticeable performance gap exists between TSOM and apg-STMD on image sequence 6 of the RIST dataset, which exemplifies the limitations of TSOM. The challenges in image sequence 6 include
small objects with low contrast against the background,
the co-directional motion of objects and background, and
presence of high-intensity objects in the background.
In the output of the SGC dendritic layer and SGC soma layer, besides the significant presence of small objects, the intersection of high and low-intensity regions in the background is also prominent in the feature maps. The reason for these issues lies in how TSOM perceives spatiotemporal information, which integrates the changes in values in the three-dimensional space over preceding and subsequent frames. In image sequence 6, the high contrast between the bright and dark parts of the background house exceeds the contrast between the small object and its surrounding background. Consequently, a point undergoing high contrast in the temporal dimension yields a higher calculated value. This effect further propagates to the output of the Rt layer, interfering with the detection of small objects against the background. It should be noted that apg-STMD achieves better results in image sequence 6 because it incorporates additional prior information about background motion when determining whether an object is a object. In low contrast between small objects and their surrounding background, the extraction of small objects is susceptible to interference from high-contrast regions in the background, leading to decreased accuracy. 

The exceptional performance of the TSOM in detecting small objects is inspired by the avian visual circuit's remarkable sensitivity to small object motion. The RGCs on the retina possess small receptive fields, allowing moving small objects to activate thousands of RGCs along their paths sequentially. Each RGC generates a short spike sequence, rapidly activating hundreds of spatially overlapping, large receptive field neurons in the SGCs. The receptive fields of SGCs exhibit relatively small excitatory centers surrounded by larger inhibitory surrounds, which respond strongly to minute, high-speed stimuli. Each SGC neuron produces an irregular, long-lasting spike sequence, which significantly overlaps temporally with spike sequences from other SGC neurons. This process is temporally independent, meaning that when a moving small object appears continuously within the receptive fields of all SGC neurons at each time step, each neuron exhibits a unique and robust response. Consequently, avian vision displays remarkable sensitivity to moving small objects, and the TSOM model, inspired by this biological characteristic, excels in small object detection.

% Numbered list
% Use the style of numbering in square brackets.
% If nothing is used, default style will be taken.
%\begin{enumerate}[a)]
%\item 
%\item 
%\item 
%\end{enumerate}  

% Unnumbered list
%\begin{itemize}
%\item 
%\item 
%\item 
%\end{itemize}  

% Description list
%\begin{description}
%\item[]
%\item[] 
%\item[] 
%\end{description}  

% % Figure
% \begin{figure}[<options>]
% 	\centering
% 		\includegraphics[<options>]{}
% 	  \caption{}\label{fig1}
% \end{figure}

% \begin{table}[ht]
% \caption{}\label{tbl1}
% \begin{tabular*}{\tblwidth}{@{}LL@{}}
% \toprule
%   &  \\ % Table header row
% \midrule
%  & \\
%  & \\
%  & \\
%  & \\
% \bottomrule
% \end{tabular*}
% \end{table}

% Uncomment and use as the case may be
%\begin{theorem} 
%\end{theorem}

% Uncomment and use as the case may be
%\begin{lemma} 
%\end{lemma}

%% The Appendices part is started with the command \appendix;
%% appendix sections are then done as normal sections
%% \appendix

% To print the credit authorship contribution details
\printcredits
% \newpage
%% Loading bibliography style file
% \bibliographystyle{model1-num-names}
\bibliographystyle{cas-model2-names}
% \bibliographystyle{elsarticle-harv}
% Loading bibliography database
\bibliography{cas-refs}

% % Biography
% \bio{}
% \bibitem{ref10}
% Saleemi, I., Shah, M. (2013). Multiframe many–many point correspondence for vehicle tracking in high density wide area aerial videos. International journal of computer vision, 104, 198-219.
% % Here goes the biography details.
% \endbio

% \bio{pic1}
% % Here goes the biography details.
% \endbio
% \end{thebibliography}
\appendix
\onecolumn
\setcounter{figure}{0}
\setcounter{equation}{0}
\renewcommand{\thefigure}{B.\arabic{figure}}
\renewcommand{\theequation}{A.\arabic{equation}}
\section{Proposition Proof}\label{A}

\newproof{pf}{Proof of Proposition1}
\begin{pf}
To prove $E_{1}\leq E_{2}$, given that $\sum p_{i}=1$, it suffices to demonstrate Eq.\ref{eq6}
\begin{eqnarray}\label{eq6}
\prod_{i=1}^{N_{d}}\left ( 1-e_{i} \right )\leq\sum_{i=1}^{N_{d}}p_{i}\left ( 1-e_{i} \right )
\end{eqnarray}
As the monotonicity of the logarithmic function, Eq.\ref{eq7} can be obtained.
\begin{eqnarray}\label{eq7}
\sum_{i=1}^{N_{d}}\log \left ( 1-e_{i} \right )\leq \log \left ( \sum_{i=1}^{N_{d}}p_{i}\left ( 1-e_{i} \right ) \right )
\end{eqnarray}
As $p_{i}\in \left [ 0,1 \right ], e_{i}\in \left [ 0,1 \right ]$, it follows Eq.\ref{eq8}. 
\begin{eqnarray}\label{eq8}
\sum_{i=1}^{N_{d}}\log \left ( 1-e_{i} \right )\leq \sum_{i=1}^{N_{d}}p_{i}\log \left ( 1-e_{i} \right )
\end{eqnarray}
By Jensen's inequality, Eq.\ref{eq9} can be obtained. Q.E.D.
\begin{eqnarray}\label{eq9}
\sum_{i=1}^{N_{d}}p_{i}\log \left ( 1-e_{i} \right )\leq \log \left ( \sum_{i=1}^{N_{d}}p_{i}\left ( 1-e_{i} \right ) \right )
\end{eqnarray}
\end{pf}

\section{Response properties of small object motion in pigeon OT neurons}\label{app:Response properties of small target motion in pigeon OT neurons}
\subsection{Experimental paradigm}
We initially design stimulus paradigms for objects at different scales. Subsequently, based on the results obtained from the experiments using the first stimulus paradigm, we employ the object scale corresponding to the maximum response intensity to design stimulus paradigms with multiple motion speeds. Wang's research \citep{ref16} has experimentally verified that, in the same motion direction, continuously moving objects can elicit faster and more robust responses from OT neurons compared to discretely flickering objects. Hence, in the experimental paradigm of this section, all moving objects specifically refer to continuous motion, and further discussion on discretely flickering objects is omitted.

\begin{figure}[htbp]
\centering
\subfigure[]
{
 	\begin{minipage}[b]{.35\linewidth}
        \centering
        \includegraphics[scale=0.4]{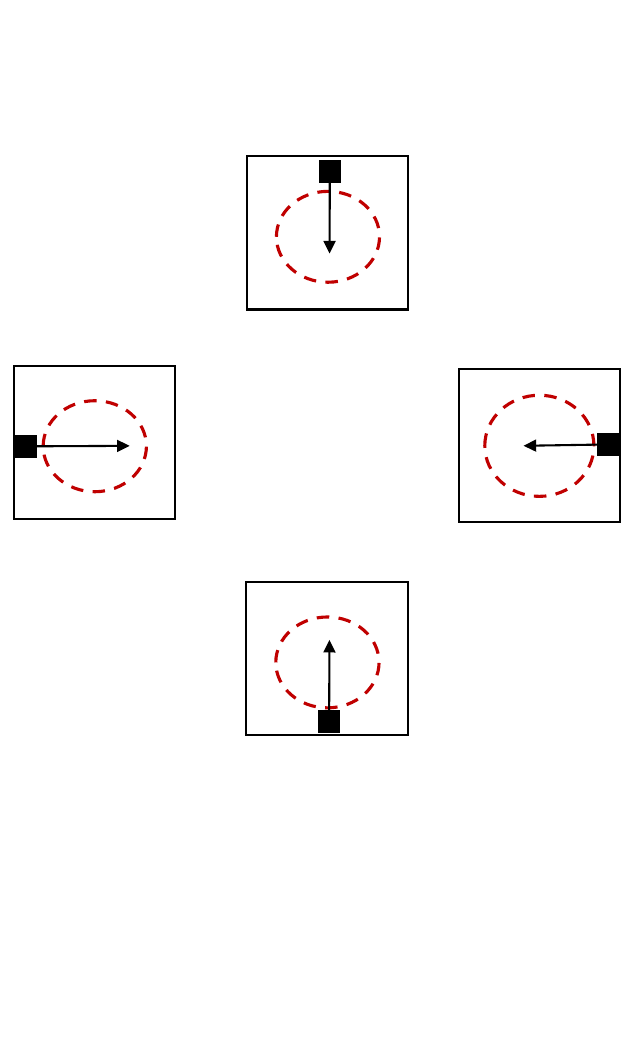}
    \end{minipage}
}	
\subfigure[]
{
 	\begin{minipage}[b]{.5\linewidth}
        \centering\hspace{-0.4cm}
        \includegraphics[scale=0.4]{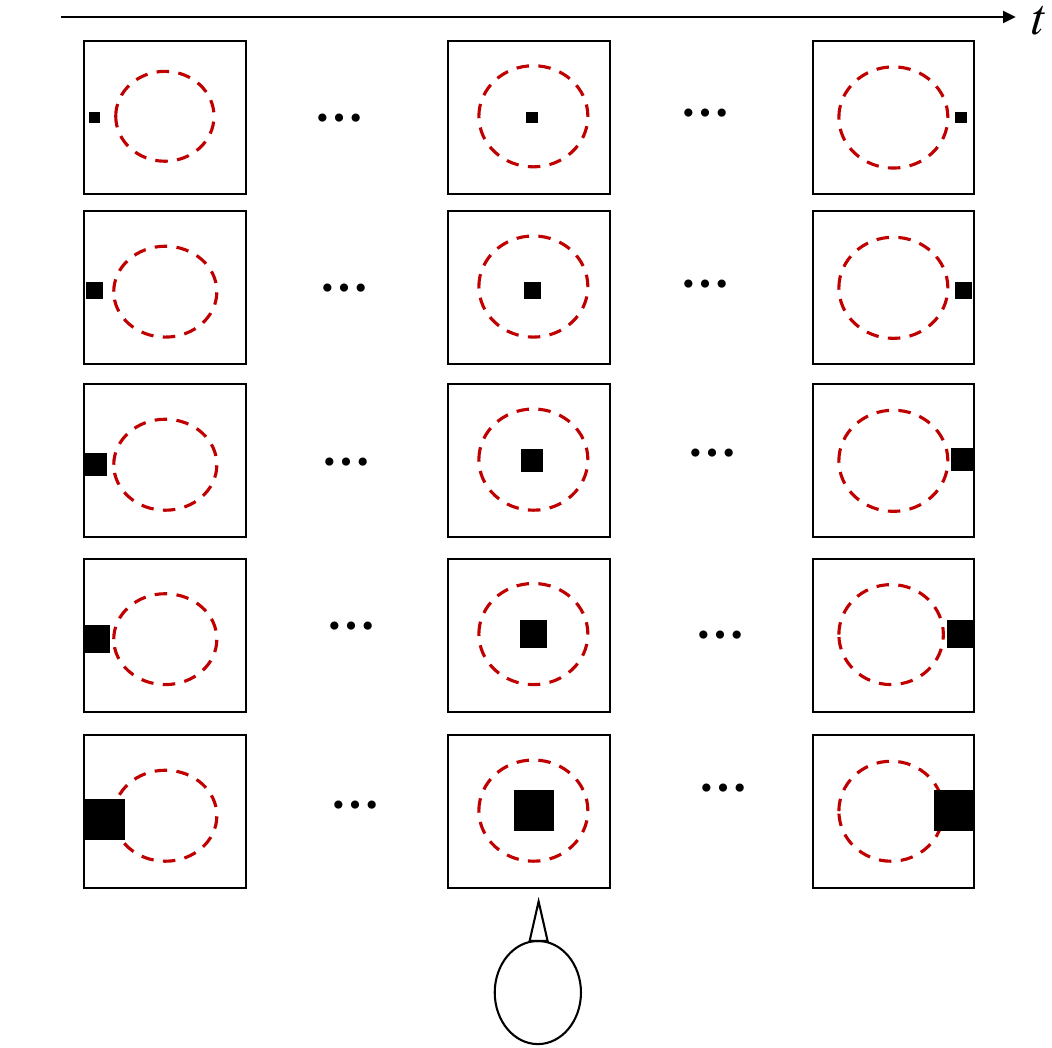}
    \end{minipage}
}
\caption{Object scale paradigms. (a) Each round comprised stimuli in four directions (0°, 90°, 180°, and 270°). Stimuli in the four directions appeared randomly, with 20 repetitions for each direction across the 20 trials. (b) The object side lengths were set to 0.36°, 0.72°, 1.8°, 3°, and 7.5°, each serving as one round of stimuli. }
\label{Sfig1}
\end{figure}

In experiments involving different object scales with a fixed motion speed, various object scale stimuli were designed to assess the preferred object scale of OT neurons, as illustrated in \cref{Sfig1}. The stimulus background was gray, with the object being a black square. The objects moved at a constant speed across the center of the receptive field, with the speed determined by adjusting the duration of the stimulus at each fixed position with a step size of 0.24°, corresponding to a motion speed of 4.8°/s.

\begin{figure}[htbp]
\centering
\subfigure[]
{
 	\begin{minipage}[b]{.35\linewidth}
        \centering
        \includegraphics[scale=0.4]{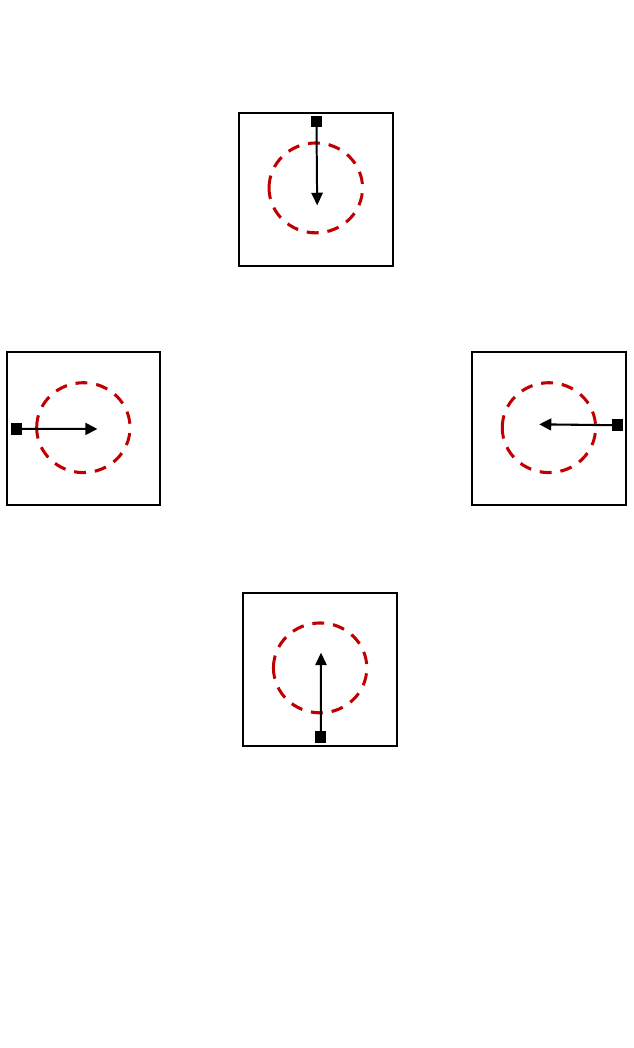}
    \end{minipage}
}	
\subfigure[]
{
 	\begin{minipage}[b]{.5\linewidth}
        \centering
        \includegraphics[scale=0.4]{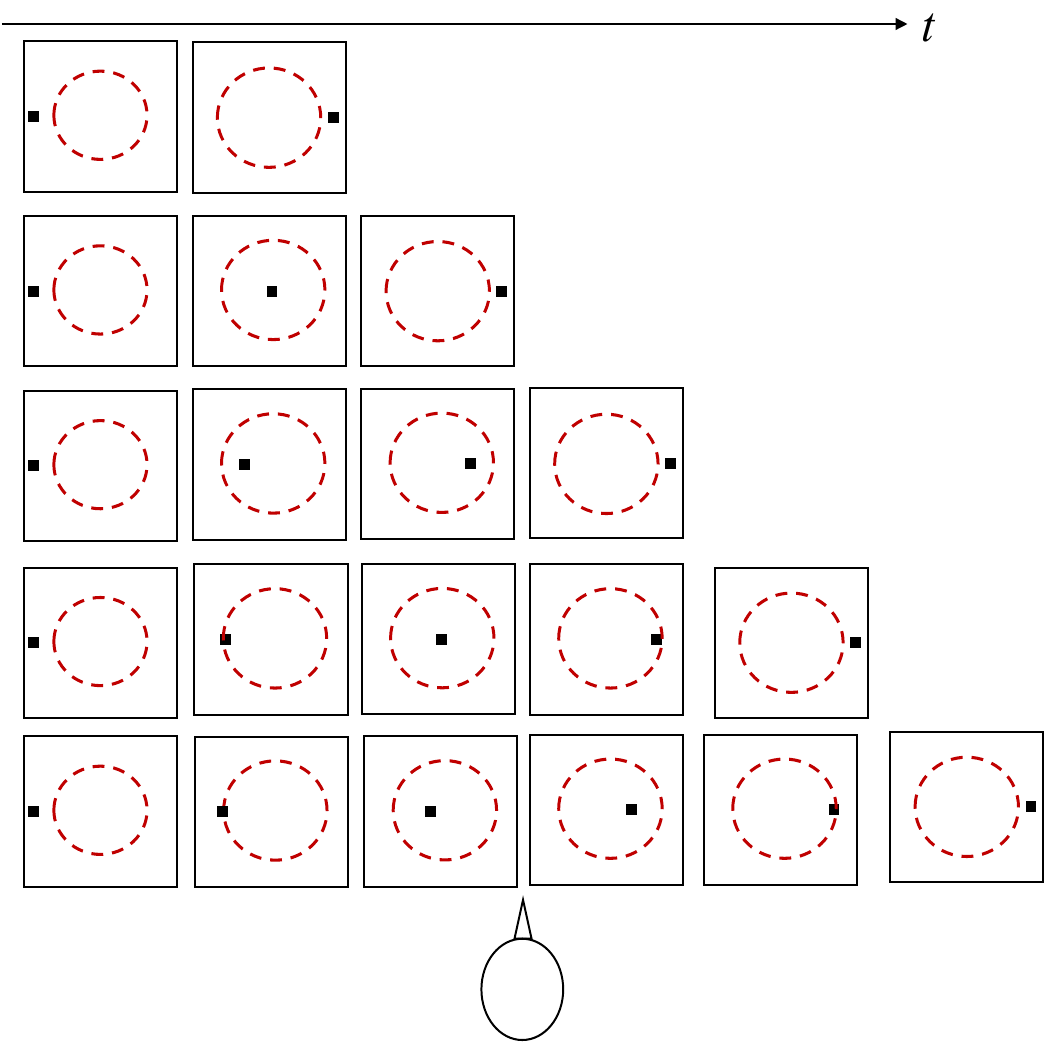}
    \end{minipage}
}
\caption{Object motion velocity paradigms. (a) Each round included stimuli in four directions (0°, 90°, 180°, and 270°). Stimuli in the four directions appeared randomly, with 20 repetitions for each direction across the 20 trials. (b) The stimulus duration was set to 10ms, 30ms, 50ms, 70ms, and 100ms, corresponding to motion speeds of 24°/s, 8°/s, 4.8°/s, 3.4°/s, and 2.4°/s, respectively. Each speed constituted one round of stimuli.
 }
\label{Sfig2}
\end{figure}

In experiments involving different motion speeds with a fixed object scale, various object speeds were designed to assess the preferred object speed of OT neurons, as illustrated in \cref{Sfig2}. The stimulus and object colors remained constant, with the object side length set to the object scale that elicited the most robust OT response in the previous experiment. The motion path maintained a constant step size of 0.24° along the trajectory.

\subsection{Neurophysiological Signal Acquisition}
\subsubsection{Microelectrode Array Implantation Surgery}
We select adult pigeons (Columba livia) weighing between 300 to 400 grams. Before the experiment, the pigeons are housed in the laboratory animal facility for at least two weeks, with free access to water and food. Before surgery, we select pigeons with good visual function and overall health and record their weights. The surgical procedure for animal implantation is as follows:
\begin{enumerate}[1)]
\item Anesthesia: Anesthesia is administered to the experimental animals via intraperitoneal injection of urethane at a ratio of 1 milliliter per 100 grams of body weight. We monitor the animals until they enter a state of anesthesia.
\item Placement in Stereotaxic Apparatus: Once the experimental animals are under anesthesia, the feathers around the ears and the region near the object brain area are trimmed. The ears are exposed, positioned at a 45° angle, and fixed in the stereotaxic apparatus, securing the animal's head.
\item Removal of Skull Bone over Object Brain Area: After disinfecting the de-feathered scalp, we make an incision with scissors and clear subcutaneous tissue and blood. A hole is drilled in the object area using a skull drill. Following the drilling, forceps are used to remove the dura mater, exposing the object region.
\item Electrode Implantation: The microelectrode array used in the surgery is made of a platinum-iridium alloy arranged in a 4x4 pattern with a spacing of 250 µm between adjacent electrodes. The length of the electrodes is 7 mm, and the tip diameter is 50 µm, with impedance ranging from 20$k\Omega$ to 50$k\Omega$. The electrode is fixed on the support of the stereotaxic apparatus and gradually lowered until its tip almost touches the object brain area, set as the depth origin. During this step, the depth indicator of the stereotaxic apparatus is zeroed. Subsequently, based on the reading of the depth indicator, the electrode is lowered to a position 800-1200$\mu m$ from the origin. After implantation, the grounding wire of the electrode is buried beneath the brain cortex and grounded. As depicted in Figure 2.3, after a recovery period of 30 to 60 minutes, when the signals stabilize, experimental stimuli and signal acquisition can commence. This procedure ensures accurate electrode implantation and stable signal recording.
\end{enumerate}
\subsubsection{Collection and Preprocessing of OT Neuron Signals}
This experiment employs a dual-screen setup featuring a control screen and a stimulus screen. The control screen is positioned externally to the experimental setup, while the stimulus screen is located within the experimental apparatus. The stimulus screen remains synchronized with the control screen, directly presenting visual stimuli to the pigeons. The stimulus screen's display resolution is 3840×2160 pixels, with a screen refresh rate of 100Hz.

The experimental paradigm is designed and implemented using Matlab and the widely used neuroscience toolbox, Psychtoolbox. Visual stimuli, located 40cm from one eye of the pigeon while the other eye is covered with tape, are presented on the display screen. The stimuli play within a region of 1080×1080 pixels on the monitor during the experiment.

The collection of neuron signals in this experiment utilizes the Cerebus™ system manufactured by the American company BLACKROCK. The Cerebus™ neural signal processor allows for real-time acquisition and analysis of action potentials (spikes), field potentials, and other physiological signals.

The signals collected by the Cerebus™ system can be categorized into two main components based on frequency: high-frequency signals and low-frequency signals. High-frequency signals refer to action potentials (spikes), which constitute the portion of the original signal above 500Hz. As the primary focus of this study lies on the high-frequency action potentials of neurons, this paper employs threshold separation on the high-frequency spikes.
\subsubsection{Receptive Fields of OT Neurons}
\begin{figure}[htbp]
\centering
\includegraphics[scale=1]{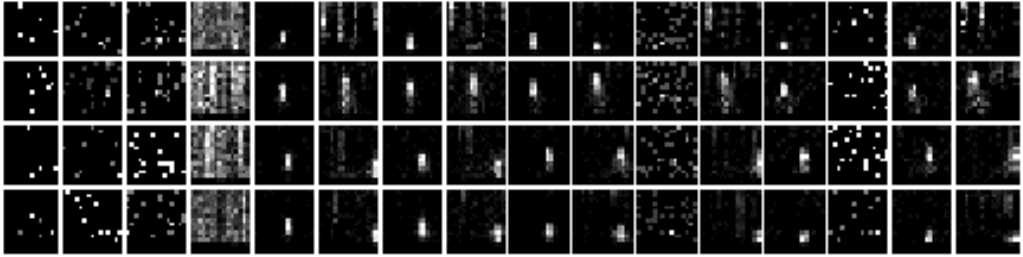}
\caption{Motion receptive field. The collection depth is 1000$\mu m$, with four rows representing four motion directions. Each column depicts the response of a specific channel.
 }
\label{Sfig3}
\end{figure}

In this section, the Spike-Triggered Average (STA) method is employed to determine classical receptive fields. The steps for determining classical receptive fields are outlined as follows:
\begin{enumerate}[1)]
\item A black square with a side length of 0.5° is moved along the four directions (up, down, left, and right) in a pseudorandom sequence. This motion is repeated 10 times for each direction, and the number of spike firings is recorded for each trial within a statistical time window. The statistical time window corresponds to the motion duration of the object in each direction.
\item For each direction, after 10 repetitions, the average value of neuron responses at the same time window and position is calculated. The neuron responses for different time windows are arranged chronologically, forming a three-dimensional response matrix containing spatiotemporal information. This matrix reflects the average activity levels of neurons under different temporal and spatial conditions, providing a comprehensive spatiotemporal perspective for subsequent data analysis.
\item Normalization is applied to the values in the matrix, and then they are mapped to pixel values in the range of 0 to 255. This generates a motion receptive field map for a specific channel neuron, as illustrated in \cref{Sfig3}.
\end{enumerate}

The receptive field maps show that signals from channels 1, 2, 3, and 14 exhibit extreme sparsity with no precise receptive field distribution. In contrast, channels 5, 7, 9, and 10 show more pronounced receptive fields, making them suitable candidates for further experimentation in the subsequent steps.
\subsection{Response Characteristics of OT Neurons to Small Object Motion}\label{Response Characteristics of OT Neurons to Small Object Motion}
In this section, spike count histograms are employed to visually depict the extent of neuronal responses. Quantification is achieved by using neuronal response firing rate (FR), defined as the average number of neuron spikes within n fixed time window bins. In this chapter's experimental context, 9 experimental subjects were utilized, comprising 127 effective recording sites, with 36 sites eligible for analysis. The subsequent discussion assumes a neuronal receptive field size of 12°×12°.
\subsubsection{Analysis of OT Neuronal Responses to Small-Scale Objects}\label{Small-Scale Targets}
\begin{figure}[htbp]
\centering
\subfigure[]
{\label{S4.a}
 	\begin{minipage}[b]{.3\linewidth}
        \centering
        \includegraphics[scale=0.65]{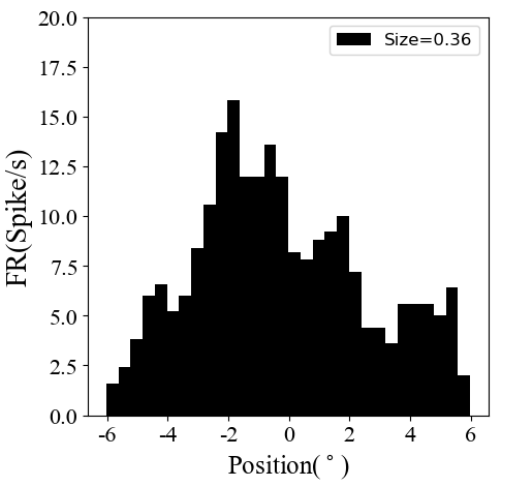}
    \end{minipage}
}	
\subfigure[]
{\label{S4.b}
 	\begin{minipage}[b]{.3\linewidth}
        \centering
        \includegraphics[scale=0.65]{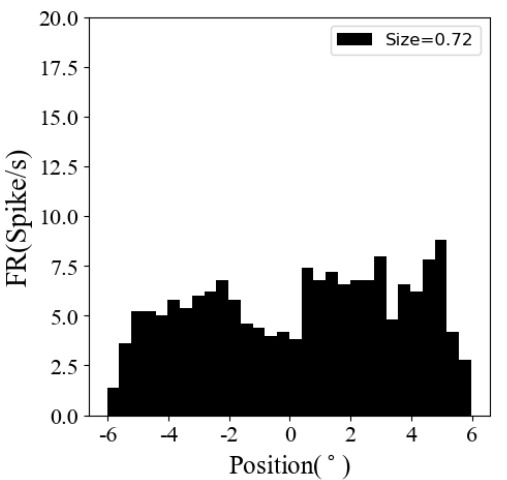}
    \end{minipage}
}
\subfigure[]
{\label{S4.c}
 	\begin{minipage}[b]{.3\linewidth}
        \centering
        \includegraphics[scale=0.65]{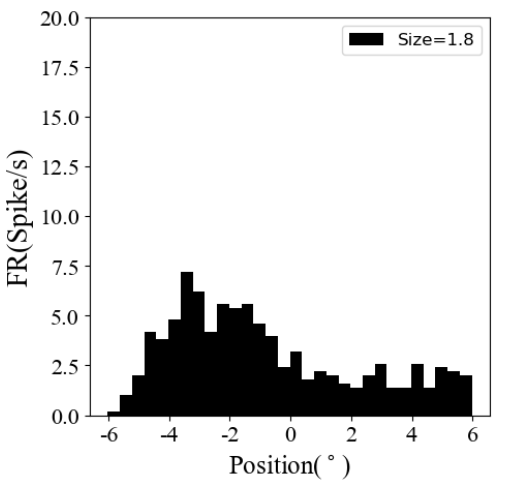}
    \end{minipage}
}	
\subfigure[]
{\label{S4.d}
 	\begin{minipage}[b]{.3\linewidth}
        \centering
        \includegraphics[scale=0.65]{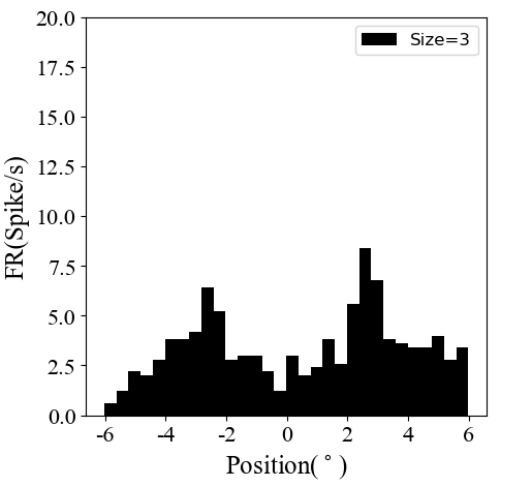}
    \end{minipage}
}	
\subfigure[]
{\label{S4.e}
 	\begin{minipage}[b]{.3\linewidth}
        \centering
        \includegraphics[scale=0.65]{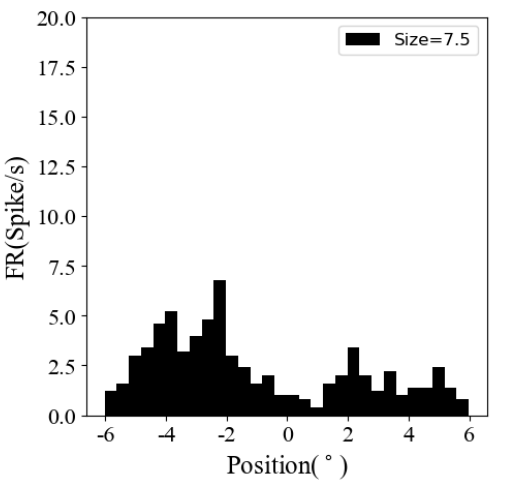}
    \end{minipage}
}	
\subfigure[]
{\label{S4.f}
 	\begin{minipage}[b]{.3\linewidth}
        \centering
        \includegraphics[scale=0.65]{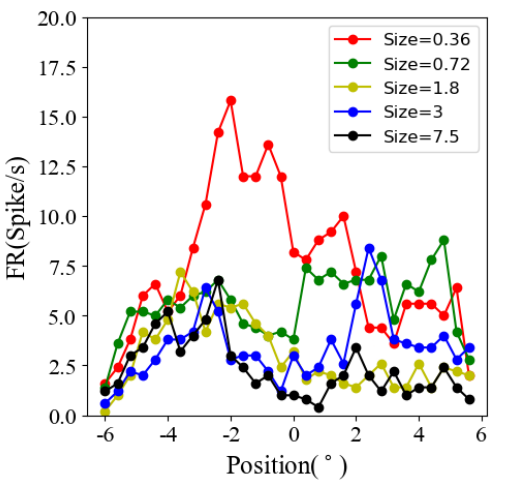}
    \end{minipage}
}	
\caption{Stimulus-response maps at different object scales. (a) Scale=0.36°, (b) scale=0.72°, (c) scale=1.8°, (d) scale=3°, (e) scale=7.5°. (f) Comparison of the responses of OT neurons.}
\label{Sfig4}
\end{figure}

The experimental results, as depicted in \cref{Sfig4} with an observation time window (bin) set at 1, reveal the following trends: from (a) to (e), it is evident that under a receptive field size of 12°×12°, the responses of OT neurons gradually decrease as the object scale increases. The smallest scale stimulus elicits the most robust response from the neurons. As the object size increases from 0.36° to 0.72°, the response significantly weakens, and as the scale further increases, the differences in responses gradually diminish. This indicates a preference of OT neurons for responding to smaller-scale objects. Beyond their preferred scale, the response intensity decreases and tends to be similar.

\cref{S4.f} consolidates the response curves for the five scales into one graph, providing a more intuitive display of the tuning pattern of neurons and a comparison of response intensities. With an increase in object scale, the maximum values of the response curves decrease. The Peak Firing Rate is commonly used as a metric for response intensity, thereby concurrently supporting the preference of OT neurons for smaller objects from the perspective of the maximum response rate.

The above results demonstrate that, compared to larger-scale objects, smaller-scale objects can more effectively induce strong responses in OT neurons. This experiment's object scale that induces the most robust neuronal response is 0.36°.
\subsubsection{Analysis of OT Neuronal Responses to Velocity Preferences for Small Objects}\label{Velocity Preferences for Small Targets}
\begin{figure}[tbp]
\centering
\subfigure[]
{\label{S5.a}
 	\begin{minipage}[b]{.3\linewidth}
        \centering
        \includegraphics[scale=0.65]{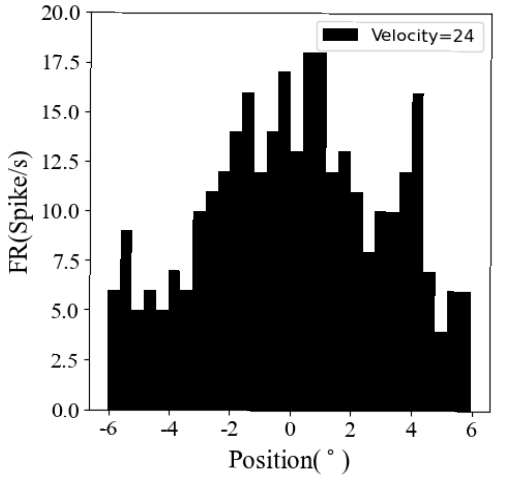}
    \end{minipage}
}	
\subfigure[]
{\label{S5.b}
 	\begin{minipage}[b]{.3\linewidth}
        \centering
        \includegraphics[scale=0.65]{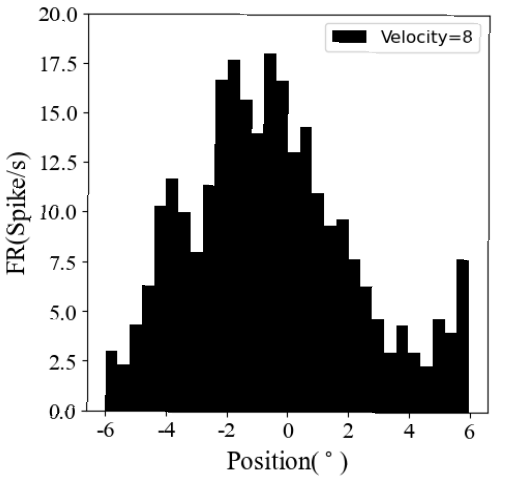}
    \end{minipage}
}
\subfigure[]
{\label{S5.c}
 	\begin{minipage}[b]{.3\linewidth}
        \centering
        \includegraphics[scale=0.65]{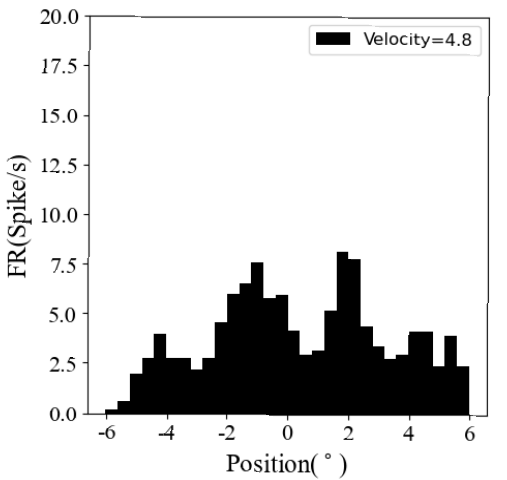}
    \end{minipage}
}	
\subfigure[]
{\label{S5.d}
 	\begin{minipage}[b]{.3\linewidth}
        \centering
        \includegraphics[scale=0.65]{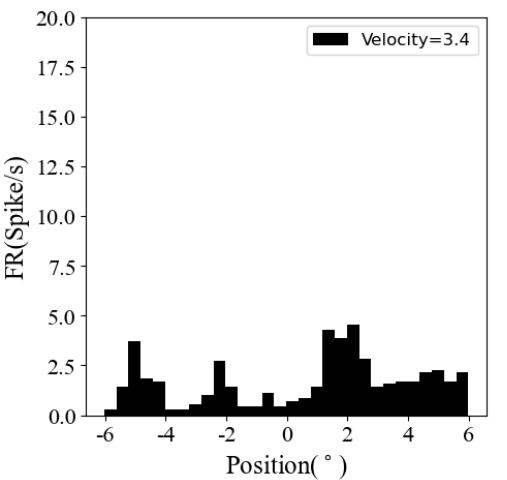}
    \end{minipage}
}	
\subfigure[]
{\label{S5.e}
 	\begin{minipage}[b]{.3\linewidth}
        \centering
        \includegraphics[scale=0.65]{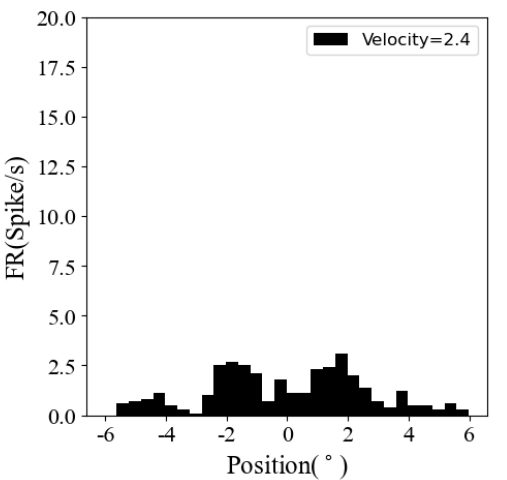}
    \end{minipage}
}	
\subfigure[]
{\label{S5.f}
 	\begin{minipage}[b]{.3\linewidth}
        \centering
        \includegraphics[scale=0.65]{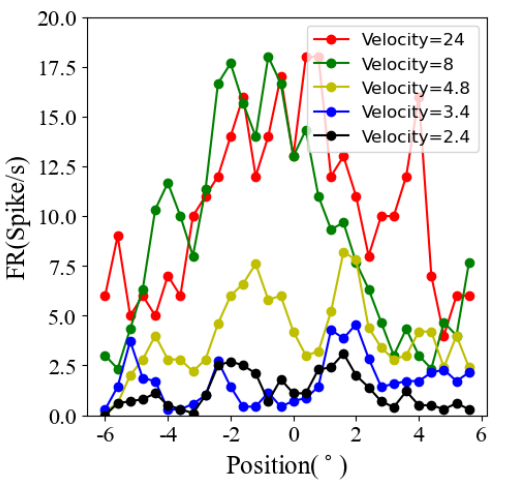}
    \end{minipage}
}	
\caption{Stimulus-response maps for different object motion speeds. (a) Velocity=24°/s, (b) Velocity=8°/s, (c) Velocity=4.8°/s, (d) Velocity=3.4°/s, (e) Velocity=2.4°/s, (f) Comparison.}
\label{Sfig5}
\end{figure}

The tuning responses of Optic Tectum (OT) neurons to small objects at different velocities, as shown in \cref{S5.a,S5.b,S5.c,S5.d,S5.e}, are analogous to the previous experimental setup, with \cref{S5.e} representing the aggregation of response curves for all scales. Significant differences in responses are observed between fast (24°/s, 8°/s, 4.8°/s) and slow (3.4°/s, 2.4°/s) speeds, particularly evident in Fig.\ref{Sfig5}(e), while the differences are less pronounced within the fast and slow categories. This is attributed to slower object movement, where, in this experiment, the object's duration in a single frame stimulus is longer, resembling discrete flickering stimuli rather than continuous motion. According to the study by Wang et al. \citep{ref16} OT neurons prefer continuous motion, hence exhibiting weaker responses at slow speeds.

\end{document}